\documentclass[manuscript]{acmart}
\usepackage{colortbl}
\usepackage{xspace}
\usepackage{soul}
          
\PassOptionsToPackage{warn}{textcomp} 
\usepackage{textcomp}                
\usepackage{multirow}
\usepackage{tabu}
\usepackage{lipsum}
\usepackage{mwe}
\usepackage{mathptmx}
\usepackage{stfloats}

\usepackage{enumitem}
\usepackage{xspace}
\usepackage{mdframed}
\usepackage{hyperref}
\setlist[enumerate]{noitemsep, topsep=0pt, leftmargin=15pt}
\setlist[itemize]{noitemsep, topsep=0pt, leftmargin=15pt}
\usepackage{xcolor}

\newcommand{\praxa}{\mbox{\sc Praxa}\xspace}
\newcommand{\PSL}{\textsc{Praxa Specification Language}\xspace}
\newcommand{\psl}{\mbox{\sc PSL}\xspace}
\newcommand{\bpstart}[1]{\noindent{\textbf{#1.}}}

\definecolor{green}{RGB}{89,161,79}
\definecolor{orange}{RGB}{242,142,43}
\newcommand{\codeN}[1]{\texttt{\textcolor{green}{#1}}}
\newcommand{\codeO}[1]{\texttt{\textcolor{orange}{#1}}}
\definecolor{blue}{RGB}{78,121,167}
\definecolor{yellow}{RGB}{240,189,39}
\definecolor{pink}{RGB}{246,114,128}
\definecolor{purple}{RGB}{148,103,189}
\definecolor{customgray}{RGB}{237, 237, 237}

\usepackage{colortbl}
\definecolor{lightgray}{gray}{0.92}

\newif\ifnotes
\notestrue

\AtBeginDocument{%
  }

\setcopyright{acmlicensed}
\copyrightyear{2018}
\acmYear{2018}
\acmDOI{XXXXXXX.XXXXXXX}

\acmJournal{POMACS}
\acmVolume{37}
\acmNumber{4}
\acmArticle{111}
\acmMonth{8}

\acmSubmissionID{4133}

\begin{document}

\title{Bridging Natural Language and Interactive What-If Interfaces via LLM-Generated Declarative Specifications}

\author{Sneha Gathani}
\email{sgathani@umd.edu}
\orcid{0000-0002-0706-7166}
\affiliation{%
  \institution{University of Maryland, College Park}
  \city{College Park}
  \state{Maryland}
  \country{USA}
}

\author{Sirui Zeng}
\affiliation{%
 \institution{University of Maryland, College Park}
 \city{College Park}
 \state{Maryland}
 \country{USA}
}
\email{szeng124@umd.edu}

\author{Diya Patel}
\affiliation{%
 \institution{University of Maryland, College Park}
 \city{College Park}
 \state{Maryland}
 \country{USA}
}
\email{dpatel90@terpmail.umd.edu}

\author{Ryan Rossi}
\affiliation{%
 \institution{Adobe Research}
 \city{San Jose}
 \state{California}
 \country{USA}
}
\email{ryrossi@adobe.com}

\author{Dan Marshall}
\affiliation{%
 \institution{Microsoft Research}
 \city{Seattle}
 \state{Washington}
 \country{USA}
}
\email{dan.Marshall@microsoft.com}

\author{\c{C}a\u{g}atay~Demiralp}
\affiliation{%
  \institution{AWS AI}
  \city{New York}
  \state{New York}
  \country{USA}
}
\affiliation{%
  \institution{MIT CSAIL}
  \city{Cambridge}
  \state{Massachusetts}
  \country{USA}
}
\email{cagatay@csail.mit.edu}

\author{Steven Drucker}
\affiliation{%
 \institution{Microsoft Research}
 \city{Seattle}
 \state{Washington}
 \country{USA}
}
\email{sdrucker@microsoft.com}

\author{Zhicheng Liu}
\affiliation{%
  \institution{University of Maryland, College Park}
  \city{College Park}
  \state{Maryland}
  \country{USA}
}
\email{leozcliu@umd.edu}

\renewcommand{\shortauthors}{Gathani et al.}

\begin{abstract}
  What-if analysis (WIA) lets users explore hypothetical scenarios by adjusting parameters, applying constraints, and scoping data through interactive interfaces. 
Current tools fall short: spreadsheet and BI tools require laborious setup, while LLM-generated interfaces frequently misinterpret analytical intent.
It remains unclear whether an explicit intermediate representation is needed, or whether LLMs can generate WIA interfaces directly.
We build a benchmark of 405 WIA questions across 11 types and 5 datasets with human-authored ground-truth intent, and find that direct one-shot generation across three LLMs fails to produce a working interface in 17--45\% of cases and, when interfaces render, often misrepresents the question's objective (up to 77.5\%) and constraints (up to 58.2\%).
We therefore present a two-stage workflow that translates NL questions into \PSL (\psl) specifications, then compiles them into interfaces.
Because errors localize to named properties, taxonomy-guided repair raises correctness from 52.42\% to 80.42\%, unlike errors in directly generated code.

\end{abstract}



\received{10 September 2026}

\maketitle

\section{Introduction}
\label{sec:introduction}
What-if analysis (WIA) involves exploring and comparing multiple scenarios by dynamically adjusting parameters, applying explicit constraints, and scoping data subsets to make data-driven decisions.
For example, a business analyst interested in how marketing spend affects profit across regions might begin with a what-if question such as: \textit{What happens to Q4 profit if marketing spend increases by 15\% in the US but decreases by 10\% in Europe?}
They may then iteratively vary these parameters across regions, add new effects such as campaign spend, or invert the question to ask what strategy reaches a target profit.
Performing such multi-step analyses requires modifying assumptions, examining immediate results, and revisiting previously explored scenarios.
This in turn demands interactive interfaces that combine parameter controls (e.g., sliders, selectors, checkboxes) with linked visualizations that update in real time, making it possible to explore scenarios, adjust values fluidly, and compare results.

Current tools, however, do not adequately support this process.
Spreadsheet-based tools like Excel and business intelligence (BI) platforms (e.g., Tableau, Power BI, Salesforce Einstein) require users to manually configure parameters, bind formulas, and specify low-level analytic settings.
Conversely, LLM-generated interfaces translate natural language (NL) directly into interactive dashboards without manual setup, but are semantically fragile and frequently misinterpret analytical intent.

To understand how often and in what ways this occurs, we conduct a formative experiment on direct one-shot generation under a deliberately strong approach--a prompt iteratively hardened against incidental implementation errors unrelated to analytical understanding, then frozen prior to evaluation--over a stratified set of WIA questions drawn from our benchmark (spanning 11 WIA types and 5 datasets), each with human-authored ground-truth analytical intent (\autoref{sec:benchmark}).
Across three state-of-the-art LLMs, direct generation frequently fails: it produces no working interface in roughly 17--45\% of cases depending on model, and even when interfaces render, they often omit or misrepresent the question's stated objective (up to 77.5\% of applicable cases) and constraints (up to 58.2\%) (\autoref{sec:direct_NL_interfaces}).
Crucially, these are not loud failures: the interfaces render, respond to interaction, and look plausible, so a misrepresented objective or a dropped constraint surfaces only under close, line-by-line inspection of the underlying code.
These failures, together with those of spreadsheet and BI tools, share a common root: analytical intent has no existence independent of the artifact that implements it--dispersed across formulas and control bindings in the former, entangled in generated code in the latter--so confirming a question was captured correctly means reading that artifact line by line.

Our results therefore motivate separating analytical intent from its implementation: before generating an interface, the system should first produce an explicit representation of what the analysis is meant to do.
Declarative representations offer a promising way to achieve this separation.
Across databases~\cite{tian2024sqlucid,fu2023catsql,tian2025text,song2022voicequerysystem}, data visualization~\cite{ko2024natural,tian2024chartgpt,song2022rgvisnet,luo2021synthesizing,luo2021natural}, and UI generation~\cite{kin2012proton,heer2023living}, grammars and declarative specifications have been used as intermediate representations that capture high-level intent while abstracting away lower-level implementation details.
WIA is amenable to a similar formulation because its analytical logic can be expressed compositionally through interconnected primitives.

Recent work by Gathani et al.~\cite{gathani2026praxa} formalizes this structure through \praxa, a compositional grammar for WIA, and \PSL (\psl), a declarative specification language that encodes it.
That work establishes the grammar, demonstrates its expressiveness, and illustrates its use as an intermediate representation, but does not establish whether that mediation is \textit{reliable}: whether LLMs can generate valid and semantically faithful specifications from NL questions, what errors arise when they do not, and whether those errors can be systematically repaired.
We take up these questions by treating \PSL as an instance of this explicit representation and building an end-to-end NL-to-interface workflow around it, in two stages (\autoref{sec:workflow}): \\
\textbf{(1) NL to Declarative Specification.} An LLM translates the NL question into a \PSL specification that explicitly represents the intended analytical logic. Because intent is represented independently of the interface implementation, errors can be identified, localized, and repaired \textit{before} interface generation--a capability our formative experiment (\autoref{sec:direct_NL_interfaces}) shows is largely absent when models produce interface code directly. \\
\textbf{(2) Declarative Specification to Interactive Interface.} The resulting specification is compiled into an interactive visual interface whose visualizations and parameter controls are bound to the underlying data, enabling users to dynamically explore and compare what-if scenarios. \\
Although instantiated with \PSL, this two-stage separation of intent from implementation applies more broadly to structured intermediate representations for analytical interface generation.

Using this workflow as a research probe, we ask:\\
\textbf{RQ1.} How reliably do LLMs translate NL WIA questions into \PSL specifications, and where do they fail? (\autoref{subsec:generation}) \\
\textbf{RQ2.} Does the explicit structure of \PSL enable systematic error detection and repair? (\autoref{subsec:repair}) \\
\textbf{RQ3.} How do specification errors propagate into compiled interfaces? (\autoref{sec:rq3}) \\
We answer these on the same benchmark, generating specifications with three state-of-the-art LLMs and auditing errors against human-authored ground truth (\autoref{sec:intermediate_NL_interfaces}).
Across models, half of specifications (52.42\%) are correct without intervention, and taxonomy-guided targeted repair raises this to 80.42\%.
Whereas our formative experiment found that direct generation's failures are entangled in generated code and surface only under close, line-by-line inspection, \PSL's errors are tied to named specification properties--scope, objective, constraints, among others--making them localizable and directly repairable.

In summary, we make the following contributions:
\begin{itemize}[left=0.05em]
\item A \textit{benchmark} of 405 WIA questions across 11 types and 5 datasets, annotated with human-authored ground-truth analytical intent.
\item A \textit{formative experiment} on direct LLM generation of WIA interfaces that motivates the need for an explicit intermediate representation, revealing that without one, generation fails both structurally (interfaces that do not render) and silently (interfaces that render but misrepresent intent) across three state-of-the-art models.
\item A \textit{two-stage workflow} that operationalizes \PSL as an intermediate representation, translating NL WIA questions into interactive interfaces while keeping analytical intent explicit, inspectable, and repairable.
\item An \textit{error taxonomy} characterizing where and how \PSL generation fails, and \textit{a targeted repair strategy} that raises specification correctness from 52.42\% to 80.42\%.
\item An analysis of how \textit{undetected specification errors propagate into compiled interfaces}, underscoring why an explicit representation matters for reliable NL-driven WIA.
\end{itemize}
\section{Background}
\label{sec:background}
\subsection{Overview of What-If Analysis}
WIA is a multi-step form of analysis that enables exploring hypothetical scenarios by varying input parameters within an underlying model and observing resulting changes in outcome variables~\cite{gathani2026praxa}. 
The underlying model encodes relationships between input parameters (e.g., \textit{Revenue}, \textit{Cost}) and output variables (e.g., \textit{Profit}) which are functions learned via machine learning (e.g., \textit{Profit} = $f$(\textit{Revenue}, \textit{Cost}, \textit{Marketing Spend}, \textit{Customer Retention Rate})).
Users repeatedly adjust parameters, impose constraints, and focus on different data subsets to examine potential outcomes and compare scenarios.

\bpstart{Types of What-if Analysis} WIA spans several distinct analysis types as characterized in prior work~\cite{gathani2026praxa}: \textit{sensitivity analysis} examines how changes to input parameters impact outcomes; \textit{goal seeking analysis} works backward from desired outcomes to identify required inputs; and \textit{importance analysis} identifies which inputs most influence outcomes.
We discuss variants of these types in \autoref{sec:benchmark}.

\bpstart{Major Tasks in Performing WIA} Performing WIA requires several interdependent tasks: (1) selecting and integrating predictive models with data, (2) constructing scenarios by adjusting parameters and imposing constraints, (3) binding various parameters and prediction outcomes to interactive controls and visualizations that update dynamically, and (4) ensuring transparency of the underlying logic to support interpretability, error handling, and trust.
Together, these tasks demand an interface that couples parameter controls with linked visualizations that update in real time \textit{and} keeps the underlying analytical logic inspectable, so that a user can verify the analysis matches their intent.
We use these requirements--particularly model integration (1), scalable parameter and constraint control (2), and transparency of logic (4)--as the lens through which we examine existing tools.

\begin{figure*}[t]
    \centering
    \includegraphics[width=\textwidth]{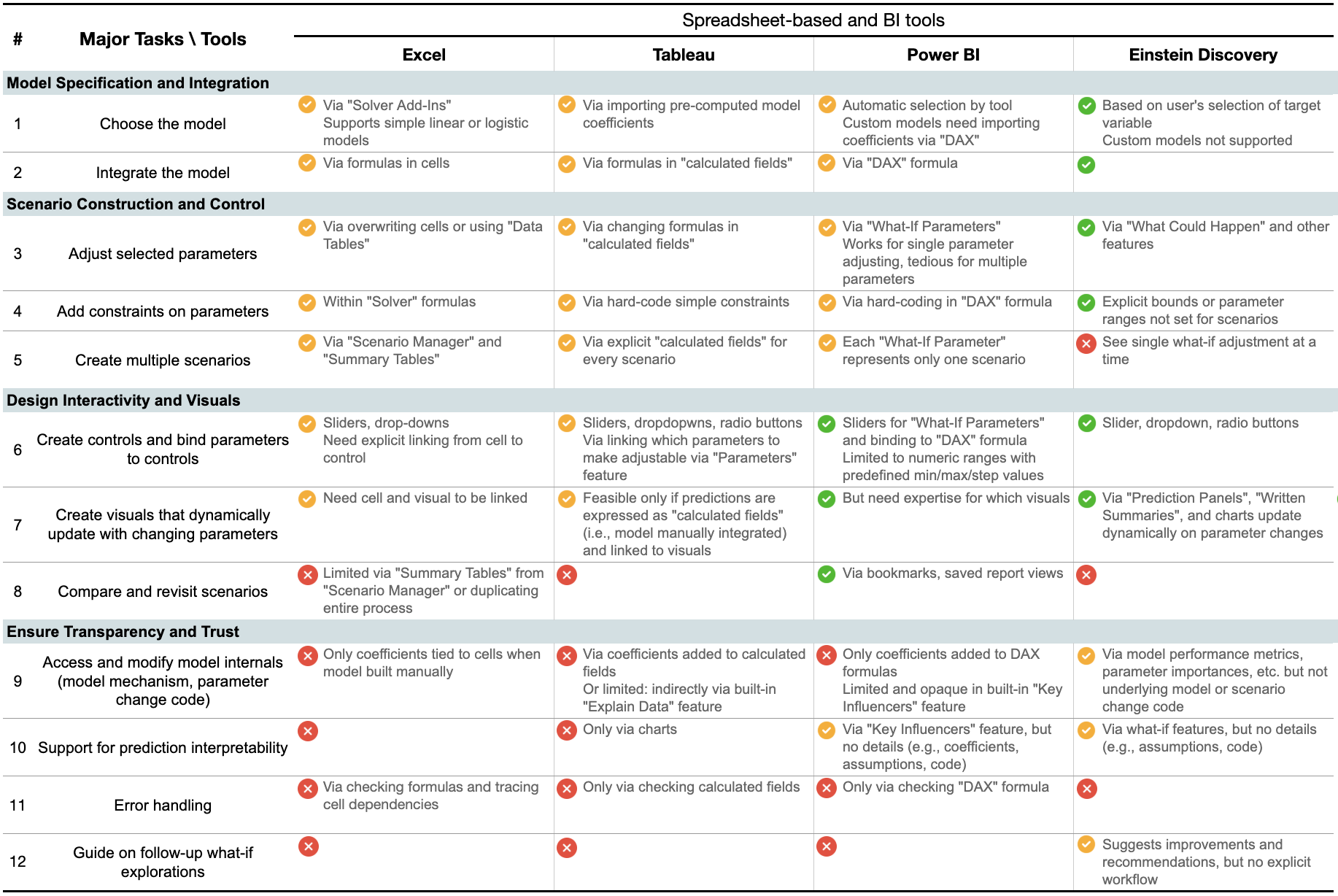}
    \caption{Summary of the authors' formative exploration of spreadsheet and BI tools (columns) across major WIA tasks (rows). Green checkmarks indicate strong support, orange indicate partial or limited support, and red crosses indicate poor or no support for each WIA task.}
    \label{fig:fig_2}
    \Description{A grid table with four spreadsheet and BI tools as columns (Excel, Tableau, Power BI, Salesforce Einstein) and major WIA tasks as rows, grouped into model specification, scenario construction, design interactivity, and transparency. Each cell holds a green check (strong support), an orange circle (partial support), or a red cross (poor or no support). Green cells are sparse and concentrated in basic parameter tasks; orange and red dominate the model-integration, constraint, transparency, and error-handling rows, showing that no tool strongly supports most WIA tasks.}
\end{figure*}

\bpstart{Existing Tools Available for WIA} Existing tools that support some of these tasks fall into two categories: spreadsheet-based and BI platforms (e.g., Excel~\cite{excel}, Tableau~\cite{tableau}, Power BI~\cite{powerbi}, Salesforce Einstein~\cite{salesforce_einstein}) and NL-to-dashboard chatbots (e.g., GPT Data Analyst~\cite{dataanalystgpt2024}, Claude~\cite{anthropic_claude_2025}).
Despite this breadth, no prior work systematically examines how these tools support WIA or the challenges users encounter.
To understand these limitations, we conducted a formative exploration of spreadsheet and BI tools (below); we examine LLM-based generation separately and in greater depth as a formative experiment (\autoref{sec:direct_NL_interfaces}), as it is the focus of this work.

\subsection{Formative Exploration of Spreadsheet and BI Tools}
To ground our understanding of how established tools support WIA, three authors independently attempted six WIA scenarios (e.g., ``what happens to churn if estimated salary is doubled?'') across four representative spreadsheet-based and BI platforms: Excel~\cite{excel}, Tableau~\cite{tableau}, Power BI~\cite{powerbi}, and Salesforce Einstein~\cite{salesforce_einstein}.
The scenarios span the three broad WIA categories--point and range sensitivity, goal-seek and constrained goal-seek, and importance analysis--and varied parameter configurations, so that each tool was probed across the diversity of WIA questions it might be asked to support (the full scenario set is listed in supplementary material).
All explorations used the \textit{Bank Customer Attrition} dataset~\cite{bankcustomerattritiondataset}; authors imported the dataset and consulted documentation and tutorials to explore each tool's WIA capabilities.
Each author invested approximately 4.5 hours across all tools, and sessions were recorded and reviewed.

\subsection{Findings}
Across all four tools, we observed four recurring challenges, summarized in \autoref{fig:fig_2}.

\bpstart{C1. Steep learning curve for model integration}
Integrating predictive models was cumbersome across all tools.
Excel's \textit{Solver Add-Ins} supported only simple models and required technical expertise.
Tableau and Power BI required manually encoding model coefficients as calculated fields or DAX expressions, making model setup error-prone and difficult to iterate (e.g., translating logistic regression coefficients into nested IF statements and binding them to data columns).
While Power BI (\textit{Key Influencers}) and Einstein Discovery provided automated modeling, they did not support custom models.

\bpstart{C2. Limited support for parameter control}
Although tools support parameter adjustments (e.g., tables, calculated fields, sliders), each scenario required manual setup and explicit formula binding.
This was repetitive and difficult to scale across multiple parameters or constraints; for example, enforcing constraints or coordinating multiple variables required complex formulas.
Einstein Discovery provided limited support for bounds but not custom constraints.

\bpstart{C3. Poor transparency of model internals}
Understanding models required tracing formulas across cells or fields, which was labor-intensive and impractical in complex multi-variable analyses.
Automated features (e.g., \textit{Key Influencers}, \textit{Top Predictors}) acted as black boxes with limited visibility or control over variables and constraints, making validation of whether the analysis matched user intent essentially infeasible.

\bpstart{C4. Limited error detection and debugging}
All tools relied on users to detect and fix errors.
Excel required manual debugging of formulas, while Tableau and Power BI produced cryptic errors (e.g., ``Cannot mix aggregate and non-aggregate arguments'') with no explanation or repair guidance.
No tool provided systematic validation of whether the configured analysis matched the user's intent.
\section{Benchmark}
\label{sec:benchmark}
Both our formative experiment on direct generation (\autoref{sec:direct_NL_interfaces}) and our evaluation of the intermediate representation-based workflow (\autoref{sec:intermediate_NL_interfaces}) draw on a common benchmark of 405 WIA questions across 11 WIA types and 5 datasets, each annotated with human-authored ground-truth analytical intent.
Building it once as shared infrastructure lets both be evaluated against the same questions and the same ground truth, on equal footing.
We release the full benchmark--questions and ground-truth intent--as supplementary materials~\cite{supplementary}.

\subsection{Datasets}
We selected five publicly available datasets (from Kaggle and the UCI ML Repository) representing distinct yet commonly encountered decision-making contexts where WIA is valuable: \textit{Bank Customer Attrition}~\cite{bankcustomerattritiondataset} (D1; customer churn, outcome \textit{Exited}), \textit{Email Campaign Management}~\cite{emailcampaigndataset} (D2; email open/ignore, outcome \textit{Email Status}), \textit{Media Spend and Sales Attribution}~\cite{mediaspendsdataset} (D3; channel spend reallocation, outcome \textit{Sales}), \textit{Marketing Campaign Response}~\cite{marketinganalyticsdataset} (D4; campaign conversion, outcome \textit{Response}), and \textit{Spotify Revenue, Expenses and Its Premium}~\cite{spotifyrevenuedataset} (D5; business financials, outcomes such as \textit{Total Revenue}).
Together they span customer, campaign, financial, and operational decision-making, yielding varied parameters and outcome variables across WIA types.
Full per-dataset feature descriptions are in supplementary material.

\subsection{What-If Analysis Types}
\begin{figure*}[t]
    \centering
    \includegraphics[width=\textwidth]{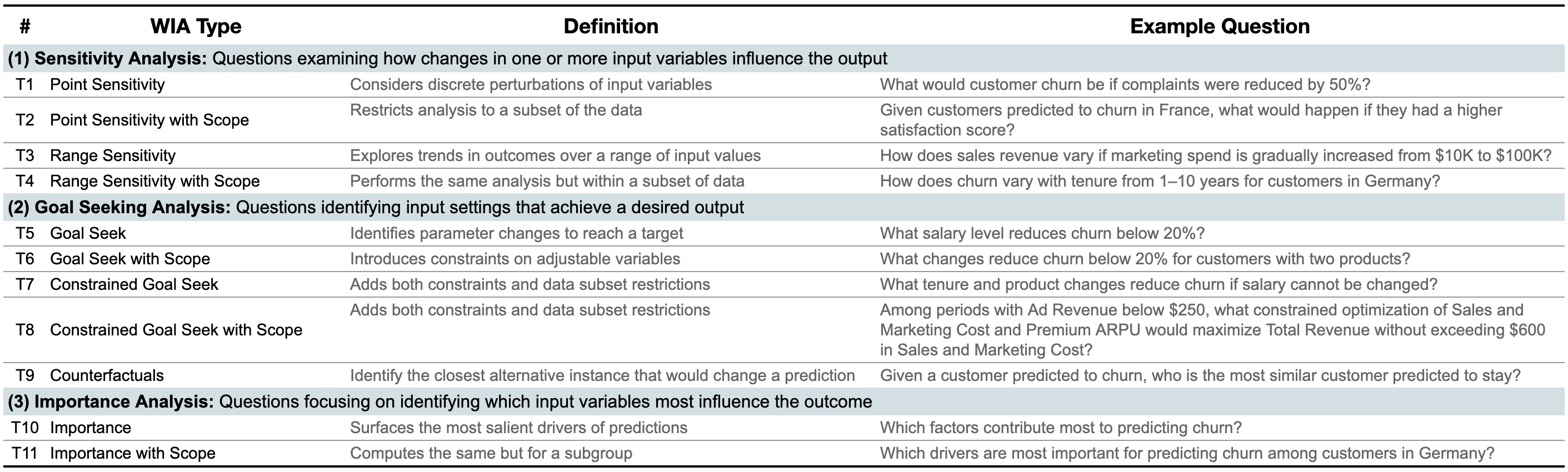}
    \caption{Eleven WIA subtypes in our benchmark grouped under three broad categories, with definitions and example questions from the Bank Customer Attrition dataset.}
    \label{fig:4_WIA_types}
    \Description{A table listing eleven WIA subtypes grouped under three categories: sensitivity analysis (point sensitivity, point sensitivity with scope, range sensitivity, range sensitivity with scope), goal-seeking (goal seek, goal seek with scope, constrained goal seek, constrained goal seek with scope, counterfactuals), and importance (importance, importance with scope). Each row gives the subtype name, a one-line definition, and an example question drawn from the Bank Customer Attrition dataset.}
\end{figure*}

Building on the three WIA categories (\autoref{sec:background}), our benchmark spans 11 subtypes varying by scope, variables, and constraints.
We illustrate these in \autoref{fig:4_WIA_types} using the \textit{Bank Customer Attrition} dataset.
Together, they help the benchmark express the diversity of WIA questions encountered in practice.

\subsection{Benchmark Construction}
\begin{figure}[!b]
    \centering
    \includegraphics[width=\columnwidth]{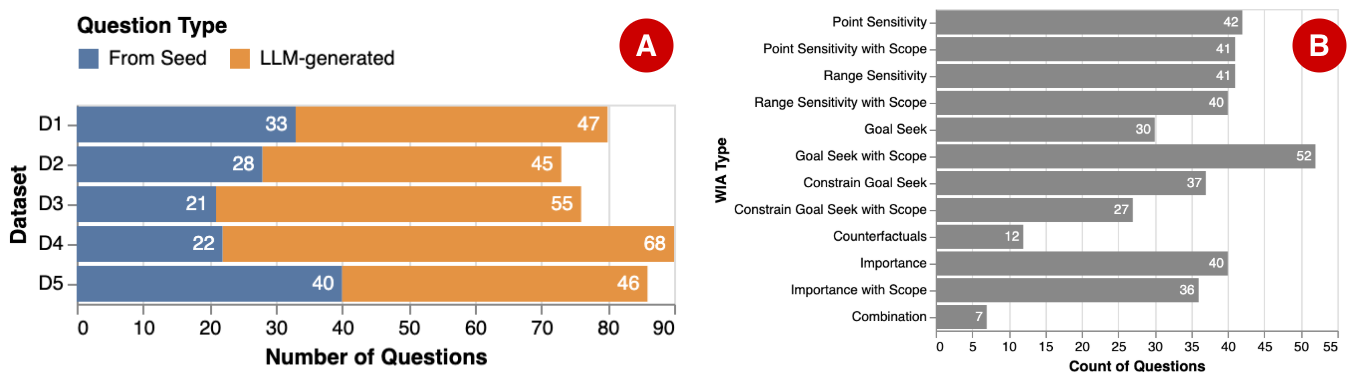}
    \caption{Distribution of the 405-question benchmark by generation method (manually authored vs.\ LLM-generated) across datasets (A) and WIA subtypes (B).}
    \label{fig:5_distribution}
    \Description{Two bar charts. Panel A shows, for each of the five datasets D1 to D5, the number of benchmark questions split into manually authored and LLM-generated portions; totals per dataset range from roughly 60 to 90 questions, with LLM-generated questions forming slightly more than half of each bar. Panel B shows the number of questions per WIA subtype, spanning about 7 to 52 questions per type, indicating coverage across all eleven subtypes with some types more represented than others.}
\end{figure}

Three coders constructed the benchmark in three steps.
First, coders adapted 152 seed questions from prior work~\cite{gathani2026praxa} to each dataset (e.g., ``What would have to change so that person X would get the loan?'' became ``What would have to change so that customer X would stay with the bank?'') across all 11 WIA types, yielding 181 hand-authored questions (44.69\%).
Second, GPT-4o scaled beyond manual authoring, generating 224 additional questions (55.31\%) with diverse phrasings and variable combinations grounded in dataset context.
Third, coders filtered the generated questions for validity, removed redundancies, and ensured balanced coverage across types and datasets, so that every WIA type is represented in every dataset where it is applicable.
The final distribution across datasets and types is shown in \autoref{fig:5_distribution}.

To verify construction quality, we confirmed that hand-authored questions preserved the structure of their seed questions (SentenceTransformer similarity 0.63--0.79 across datasets) while LLM-generated questions remained diverse (self-BLEU 0.008--0.051); details are in supplementary material.

\subsection{Ground-Truth Analytical Intent}
\label{sec:groundtruth}
For each question, we establish ground-truth analytical intent decomposed into six components common to any WIA answer regardless of how it is implemented: \textit{scope} (the data subset the analysis applies to, if any); \textit{outcome variable} (the target being analyzed); \textit{perturbation} (which input(s) change, and how--by an amount, to a value, or over a range); \textit{constraints} (bounds the solution must respect); \textit{objective} (whether to minimize, maximize, or reach a target); and \textit{analysis type} (one of the 11 WIA subtypes).
Defining ground truth at this level--rather than in terms of any particular representation--is deliberate: it lets us score interfaces produced by different generation approaches against the same criteria, so that no approach is advantaged by how its ground truth is expressed.
Two coders independently annotated all 405 questions and reconciled discrepancies through discussion to reach a single ground-truth intent per question.
Ambiguous cases (e.g., ``How does attrition probability vary for customers with 1 vs.\ 3 products?'', interpretable as descriptive comparison or as sensitivity analysis) were resolved by aligning on the appropriate WIA type.
Of 405 questions, coders initially disagreed on 179 (44.2\%); nearly all were resolved through discussion, with only 19 (4.7\%) requiring a third coder.
This initial disagreement rate reflects the genuine interpretive ambiguity of NL WIA questions rather than annotation error; the reconciled single intent per question serves as ground truth, and we document the reconciliation rationale for the disagreed cases in supplementary material.
This representation-neutral intent is scored directly against generated interfaces in our formative experiment (\autoref{sec:direct_NL_interfaces}); for the workflow, it is additionally encoded in the structured specification language we introduce in \autoref{sec:workflow}.
\section{Motivating Experiment: Limitations of Direct NL-to-Interface Generation}
\label{sec:direct_NL_interfaces}
Before introducing an intermediate representation, we first ask whether one is necessary.
If an LLM can reliably translate a WIA question directly into a correct interactive interface, an additional representation would add unnecessary complexity.
We therefore examine a strong direct-generation setting to understand where this approach fails.
Our goal is not to comprehensively benchmark code-generation strategies, but to identify whether direct generation introduces failure modes that motivate separating analytical intent from WIA implementation.

\subsection{Direct-Generation Setup}
\begin{figure}[t]
    \centering
    \includegraphics[width=\columnwidth]{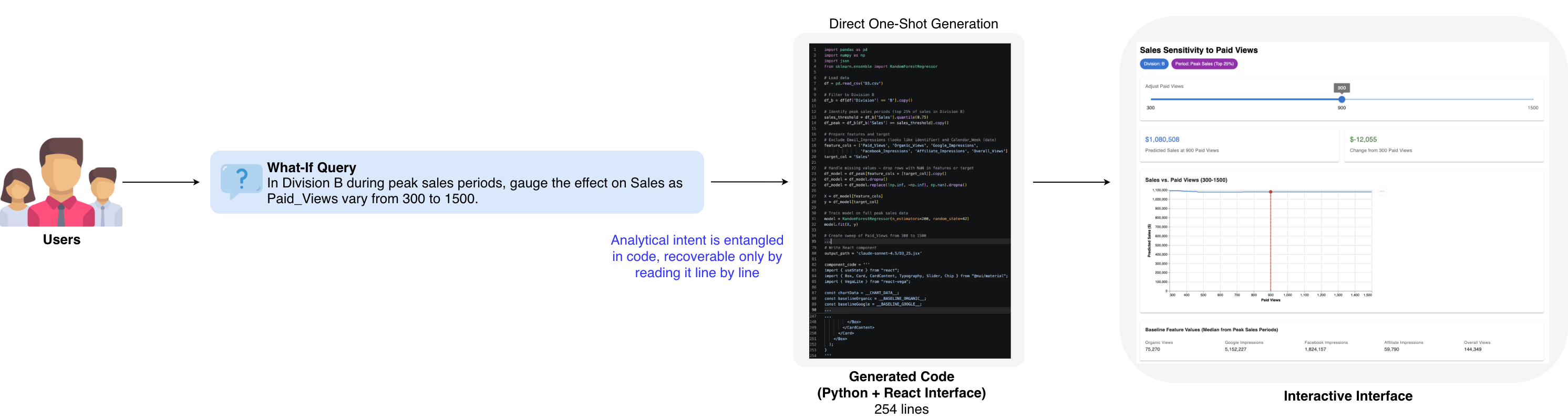}
    \caption{The direct-generation setup: an LLM receives the NL question and dataset schema and emits, in a single pass, code that both trains the model and renders the interface--with no intermediate representation at which analytical intent can be inspected. Contrast with our two-stage workflow (\autoref{fig:approach}).}
    \label{fig:direct_pipeline}
    \Description{A left-to-right pipeline diagram. On the left, a group of user icons feeds a "What-If Query" speech bubble containing a natural-language question. An arrow labeled "Direct One-Shot Generation" leads to a dark code panel showing a single generated script of about 254 lines that begins with Python (loading data, filtering to Division B, training a RandomForest) and continues into React and Vega-Lite interface code, illustrating that model logic and interface are fused in one artifact. A final arrow leads to a rendered interactive dashboard. An annotation notes that analytical intent is entangled in the code and recoverable only by reading it line by line; there is no separate intermediate representation in the flow.}
\end{figure}

For each WIA question, we provide an LLM with the NL question and the corresponding dataset schema and ask it to generate, in a single pass, the Python code that executes the analysis and implements the interactive interface end-to-end (\autoref{fig:direct_pipeline}).
The generated implementation trains the required predictive model (a scikit-learn RandomForest--a classifier for binary outcomes, a regressor for continuous ones) and produces the same three interface elements targeted by our proposed workflow: interactive parameter controls, a linked visualization, and a live prediction readout, rendered with the same stack (React, Material UI controls, and Vega-Lite charts).
Using an identical rendering target for both approaches ensures that any difference in outcome is attributable to the presence or absence of an intermediate representation, not to implementation choices--the only difference being that here the interface is authored directly by the LLM rather than compiled from any intermediate representation or specification.

We deliberately give direct generation a strong starting point.
Before evaluation, we hardened the prompt on pilot questions (disjoint from the evaluation sample) against incidental implementation pitfalls unrelated to analytical understanding--such as serialization and interface-binding errors--and then froze it, leaving it unchanged for the reported evaluation.
The complete prompt and pilot procedure are in supplementary material~\cite{supplementary}.

\bpstart{Models and sample}
We evaluate GPT-5, GPT-4o, and Claude-Sonnet-4.5\footnote{The formative experiment uses \texttt{Claude-Sonnet-4.5}; the workflow evaluation (\autoref{sec:intermediate_NL_interfaces}) uses \texttt{Claude-Sonnet-4}, a snapshot in the same model family that has since been retired by its provider, with \texttt{Claude-Sonnet-4.5} its closest available successor.
Because the two are reported separately and never compared numerically, this difference does not affect any result. 
All models used are fixed dated snapshots.} on a stratified sample of 150 questions per model, drawn proportionally across the 5 datasets and their labeled WIA types so the sample mirrors the full benchmark (\autoref{sec:benchmark}).

\bpstart{Correctness and scoring}
We examine two forms of correctness.
\textit{Execution correctness} asks whether the generated implementation produces a working interactive interface at all.
\textit{Semantic correctness} asks whether a rendered interface preserves the six representation-neutral components of analytical intent from \autoref{sec:groundtruth}--scope, outcome variable, perturbation, constraints, objective, and analysis type.
To assess semantic correctness, an automated judge receives the generated code, its execution output (or, if execution failed, the resulting error), and the ground-truth intent, and labels each of the six components as \textit{matching}, \textit{mismatching}, or \textit{not-applicable} (the last only when the ground truth has no such field for that question, e.g., an unscoped question has no scope component).
A generation is semantically correct only if no applicable component mismatches.
We validated this automated scoring against independent manual scoring of 40 samples, obtaining strong agreement (90.0\%, Cohen's $\kappa = 0.85$); full protocol and validation details are in supplementary material.

\subsection{Direct Generation Fails Both Visibly and Silently}
\begin{figure}[t]
    \centering
    \includegraphics[width=\columnwidth]{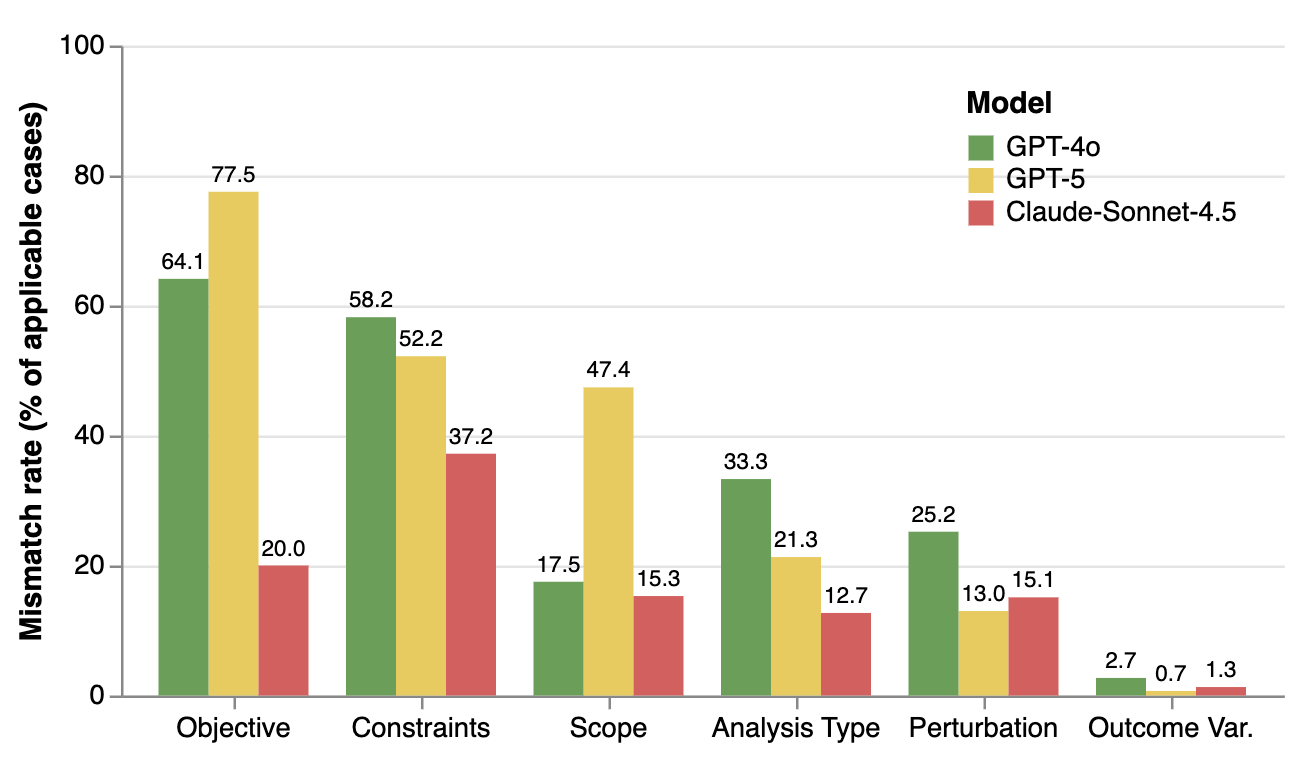}
    \caption{Per-component mismatch rates for direct generation, by model, over applicable cases. Across all three models, \textit{objective} and \textit{constraints}--the components with no forced representation in generated code--are the most frequently mismatched, while \textit{outcome variable} and \textit{analysis type} are comparatively reliable.}
    \label{fig:direct_results}
    \Description{A grouped bar chart of component mismatch rates for direct generation, with one group per intent component (scope, outcome variable, perturbation, constraints, objective, analysis type) and three bars per group for GPT-5, GPT-4o, and Claude Sonnet 4.5. Objective and constraints show the tallest bars across all three models (objective up to about 77 percent, constraints up to about 58 percent), while outcome variable and analysis type are lowest (near zero to about 33 percent), showing that the least-represented components in code are the most frequently wrong.}
\end{figure}

Direct generation exhibited two qualitatively different failure modes (\autoref{fig:direct_results}).

\bpstart{Visible failures}
Some generations did not produce a working interface at all: execution failures occurred in 20.7\% of GPT-5 generations, 44.7\% of GPT-4o generations, and 17.3\% of Claude-Sonnet-4.5 generations.
These failures are at least self-announcing--the intended interface never becomes usable, so the user knows something went wrong.

\bpstart{Silent failures}
More concerning are semantic failures, in which a rendered, interactive interface does not faithfully represent the analysis the question asked for.
Among interfaces that rendered successfully, 35.5--81.9\% (depending on model) contained at least one component mismatch relative to the ground-truth intent.
Crucially, the mismatches concentrated on the components with \textit{no forced representation in code}: for every model, \textit{objective} and \textit{constraints} were among the weakest (\autoref{fig:direct_results}), with substantial \textit{scope} errors as well.
Because these rates are computed only over questions whose ground truth specifies the component, the applicable denominators vary--constraints and objective, in particular, are specified in a subset of questions (43--55 and 80--131 of 150, respectively; full denominators in supplementary material)--so we report them as per-component mismatch rates rather than as a single aggregate.

\begin{figure*}[t]
    \centering
    \includegraphics[width=\textwidth]{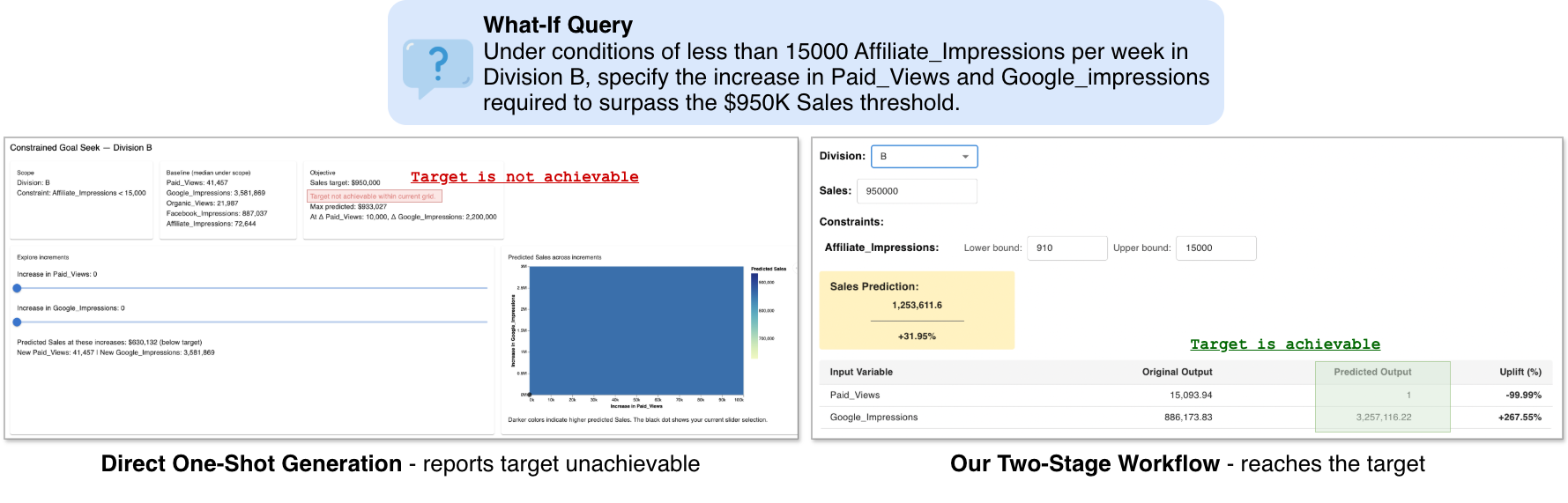}
    \caption{A directly generated interface that silently encodes a wrong analysis. For the constrained goal-seek ``\ldots specify the increase in Paid\_Views and Google\_Impressions required to surpass the \$950{,}000 Sales threshold,'' GPT-5's interface (left) renders fully but reports the target as \textit{unachievable} (max \$933{,}027): the stated scope never shapes the trained model and the search grid is too coarse to move the prediction. Compiling the same intent through our workflow (right) reaches the target, predicting \$1{,}253{,}612 with the constraint enforced. The two interfaces are visually comparable; only one is analytically correct.}
    \label{fig:direct_example}
    \Description{Two interface screenshots side by side under one what-if query about reaching a \$950,000 sales target within a scoped subset. The left interface, from direct one-shot generation, shows scope, baseline, and objective cards, sliders for two inputs, and a nearly uniform blue heatmap; it reports the target as unachievable, with a maximum predicted value of \$933,027 below the \$950,000 goal. The right interface, from the two-stage workflow, shows a division selector, a sales target field, an enforced affiliate-impressions constraint, and a sales prediction of \$1,253,612 (a 31.95 percent uplift), reaching the target. The two look comparable, but only the right one is analytically correct.}
\end{figure*}

\bpstart{Why these failures are hard to catch}
The defining property of these failures is that the interface looks right: it renders, the controls respond, and the numbers update--yet the analysis underneath is wrong, and nothing on screen reveals it.
Consider a constrained goal-seek: \textit{``Under conditions of less than 15{,}000 Affiliate\_Impressions per week in Division B, specify the increase in Paid\_Views and Google\_Impressions required to surpass the \$950{,}000 Sales threshold.''}
This asks a concrete question--by how much must two inputs rise to push predicted Sales past a target, within a specified subset of the data--and it has an answer: our workflow, compiling the same intent, finds input increases that reach the target, predicting \$1{,}253{,}612 in Sales (\autoref{fig:direct_example}, right).
GPT-5's directly generated interface (\autoref{fig:direct_example}, left) instead reports that the target is \textit{unachievable}: its highest reachable prediction is \$933{,}027, just short of \$950{,}000, and every summary card shows the goal cannot be met.
A user would reasonably conclude the business objective is impossible--when in fact it is not.

The interface gives no hint that anything is amiss.
It presents a scope card, a baseline card, sliders for both inputs, and a heatmap of predicted Sales; it is, on its face, a complete and usable goal-seek tool.
The error is entirely in the generated code, and takes two forms.
First, although the question scopes the analysis to weeks with fewer than 15{,}000 Affiliate\_Impressions, the code applies that condition only to compute baseline values--the model itself is trained on \textit{all} of Division B, so the stated scope never actually shapes the predictions.
Second, the code searches for a solution over a fixed grid of input increases that is far too coarse relative to the data: predicted Sales barely move across the entire grid, so the search terminates convinced no combination reaches the target.
Neither of these decisions is visible in the rendered interface; both are recoverable only by reading the code line by line.

This pattern--a fully functional interface that silently encodes a wrong analysis--recurs across models and questions.
On a question asking to \textit{maximize} response likelihood, two of three models instead optimized total spending, substituting the wrong outcome variable; on another, a model treated a goal-seek phrased around reaching a target rate as a plain sensitivity sweep, never optimizing at all; on a third, two of three models dropped an explicit bound limiting how far a recommended change could deviate from a customer's profile, leaving those variables unconstrained.
In every case the interface rendered and behaved plausibly, and the omitted objective, misapplied scope, or dropped constraint was visible only in the generated code.
That the same error often recurs across independently generated models on the same question indicates these are properties of the task as posed, not one model's slip: nothing in a one-shot generation contract forces a model to name and commit to the scope, objective, or constraint it is actually implementing, so any of them can be quietly mis-encoded while the interface still looks correct.
We provide additional worked examples in supplementary material.

\subsection{Why an Explicit Representation?}
These findings do not imply that direct code generation can never produce correct WIA interfaces; often it does.
The problem is that when it does not, nothing reveals it.
Across our examples, the failures were not crashes or malformed output but analytically wrong interfaces that rendered and behaved normally--a goal-seek that declared an achievable target impossible, a comparison silently re-scoped to a different set of weeks, an objective quietly dropped in favor of a plain sweep.
What these share is a single underlying cause: the meaning of the requested analysis and its implementation are produced simultaneously and embedded in the same artifact.
The decisions that determine whether the analysis is correct--which subset to scope to, what to optimize, which bounds to enforce--are never stated anywhere the user can see; they exist only as incidental choices scattered through generated code.
As a result, a misinterpreted scope, an omitted constraint, or a substituted objective becomes indistinguishable, from the interface alone, from a faithful analysis.
There is no point in the pipeline at which the intended analysis can be checked before it is acted upon.

This motivates a different workflow.
Instead of translating a WIA question directly into interface code, we first represent the intended analysis \textit{explicitly}, as a small set of named properties--scope, outcome, perturbation, constraints, objective, analysis type--that stand apart from any implementation.
Making intent a separate, inspectable artifact creates exactly the checkpoint that direct generation lacks: the analysis can be read against the original question, validated, and repaired \textit{before} it is compiled into an interactive interface, so that a scope substitution or dropped constraint is caught as a wrong property rather than discovered as an implausible result, if ever.
In the following section, we operationalize this idea using \PSL (\psl).
\section{A Two-Stage Workflow: Separating Analytical Intent from Implementation}
\label{sec:workflow}
\begin{figure*}[!b]
    \centering
    \includegraphics[width=\textwidth]{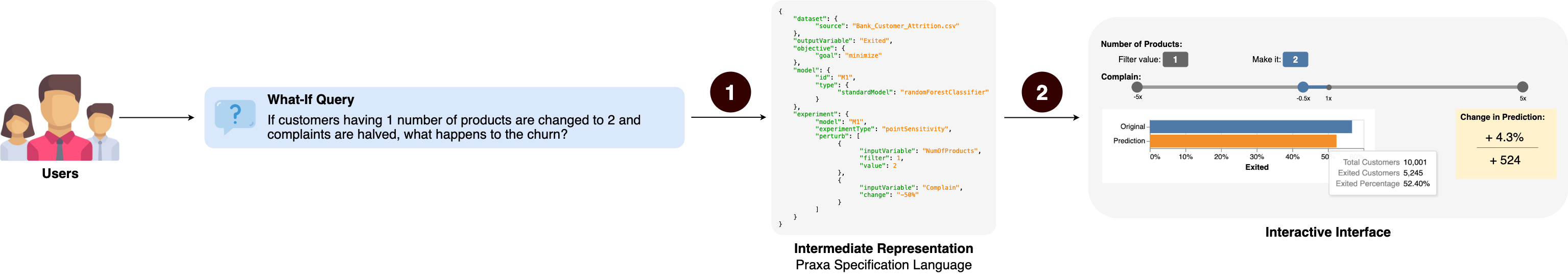}
    \caption{Our two-stage workflow: (1) translating NL WIA questions into a structured intermediate representation like \PSL (\psl), and (2) compiling it into an interactive visual WIA interface with linked controls and visualizations.}
    \label{fig:approach}
    \Description{A left-to-right pipeline diagram of the two-stage workflow. Users feed a "What-If Query" speech bubble into stage 1 (badge 1), which produces a structured PSL specification shown as JSON with named properties such as dataset, outputVariable, objective, model, and experiment. Stage 2 (badge 2) compiles that specification into an interactive dashboard on the right, with parameter controls (a products filter and a complaints slider) and a linked bar chart showing the change in predicted churn. Unlike direct generation, intent is captured explicitly in the specification before the interface is built.}
\end{figure*}

Our two-stage workflow (\autoref{fig:approach}) translates NL WIA questions into interactive visual interfaces via a declarative specification as an intermediate representation.
This directly addresses the limitations identified earlier: unlike spreadsheet and BI tools (\autoref{sec:background}), users specify analytical intent via NL rather than manual formula construction; and unlike direct LLM generation (\autoref{sec:direct_NL_interfaces}), where analysis logic is entangled with implementation, the specification separates \textit{what} to analyze from \textit{how} to implement it.
This separation links WIA components to interface controls explicitly, makes analysis logic inspectable without reading code, and provides structure for component-level validation and targeted repair rather than debugging arbitrary code.

\subsection{Stage 1: NL to Declarative Specification}
A user's NL WIA question is transformed into a structured intermediate representation that captures analytic intent.
We instantiate this representation as a JSON-based declarative specification language, \PSL (\psl), grounded in the compositional grammar of what-if analysis, \praxa~\cite{gathani2026praxa}.
\praxa models WIA workflows as compositions of three primitives: \textit{data} (variables under analysis), \textit{model} (predictive mechanisms), and \textit{interaction operations} (paired user actions and system responses, e.g., \texttt{perturb} $\rightarrow$ \texttt{rerun}, \texttt{constrain} $\rightarrow$ \texttt{optimize}).
\psl encodes these primitives as named specification properties--the same six representation-neutral components of analytical intent defined in \autoref{sec:groundtruth}--making intent explicit and inspectable against the original question rather than embedded in code.
Generation details are provided in \autoref{sec:intermediate_NL_interfaces}.

\begin{figure}[!b]
\centering
\includegraphics[width=\columnwidth]{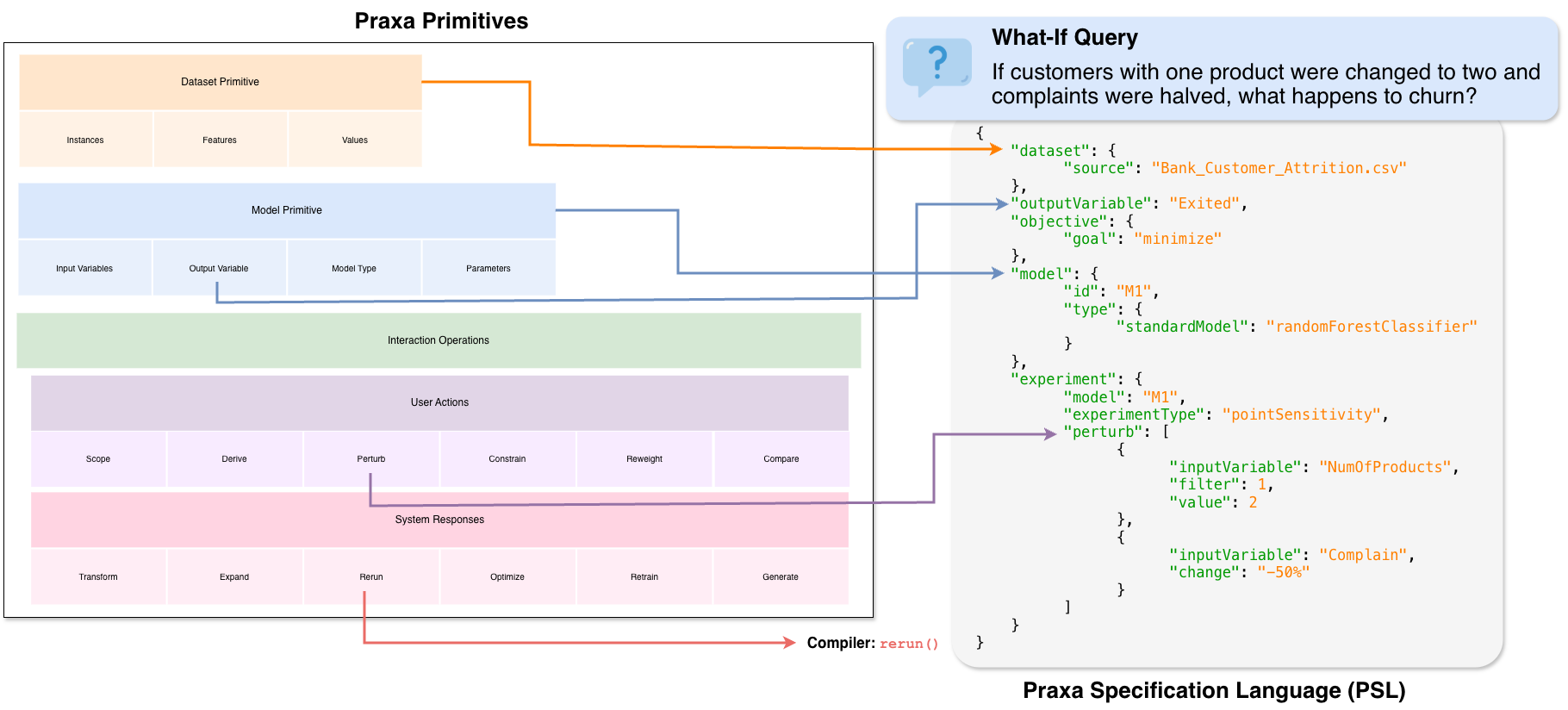}
\caption{Example \psl specification for a point sensitivity question, with properties mapping to \praxa primitives~\cite{gathani2026praxa}.}
\label{fig:3_example_psl}
\Description{An annotated example PSL specification for a point-sensitivity question, shown as JSON on the right and mapped to Praxa primitives on the left. Labeled properties include dataset, outputVariable (Exited), objective (minimize), model (a randomForestClassifier), scope, and an experiment block of type pointSensitivity with a perturb array changing input variables. Lines connect each JSON property to its corresponding data, model, or interaction-operation primitive.}
\end{figure}

\bpstart{Intermediate Representation: \psl}
We summarize key \psl properties using an example WIA question: \textit{``If customers with one product were changed to two and complaints were halved, what happens to churn?''} (\autoref{fig:3_example_psl}).

\hspace{-2mm} \codeN{dataset} encodes the data primitive, specifying the dataset over which WIA is performed (e.g., the \textit{Bank Customer Attrition} dataset for D1).

\hspace{-2mm} \codeN{outputVariable} encodes the target variable of interest, e.g., mapping `churn' to the \textit{Exited} parameter.

\hspace{-2mm} \codeN{objective} encodes the intended goal for the \codeN{outputVariable}, e.g., \codeO{minimize} churn; other goals include \codeO{maximize} or \codeO{setTarget}.

\hspace{-2mm} \codeN{model} encodes the predictive model connecting input variables to the output variable.
Each entry specifies an \codeN{id} and \codeN{type}, chosen according to the type of the \codeN{outputVariable}.
In the example, because \textit{Exited} is binary, a \codeO{randomForestClassifier} is used, though an alternative such as \codeO{logisticRegressor} is also valid.

\hspace{-2mm} \codeN{scope} (optional) defines filters restricting analysis to subsets of the \codeN{dataset} (e.g., filtering by region), supporting scoped WIA types.

\hspace{-2mm} \codeN{experiment} encodes the WIA workflow through interaction operations, specifying an \codeN{experimentType} (one of the 11 types in \autoref{fig:4_WIA_types}) along with an optional \codeN{scope} and model.
For example, \codeO{pointSensitivity} defines perturbations via \codeN{perturb} (e.g., changing variable values or percentages, with optional filters).
Other types use properties such as \codeN{top} (importance) and \codeN{constraints} (goal seek).
System responses (e.g., \codeN{rerun}, \codeN{optimize}) follow the interaction pairing defined in \praxa~\cite{gathani2026praxa}.
Multiple experiments can be composed within a single specification.
Full schema details are in supplementary material.

\subsection{Stage 2: Declarative Specification to Interactive Visual Interface}
\begin{figure*}[t]
    \centering
    \includegraphics[width=\textwidth]{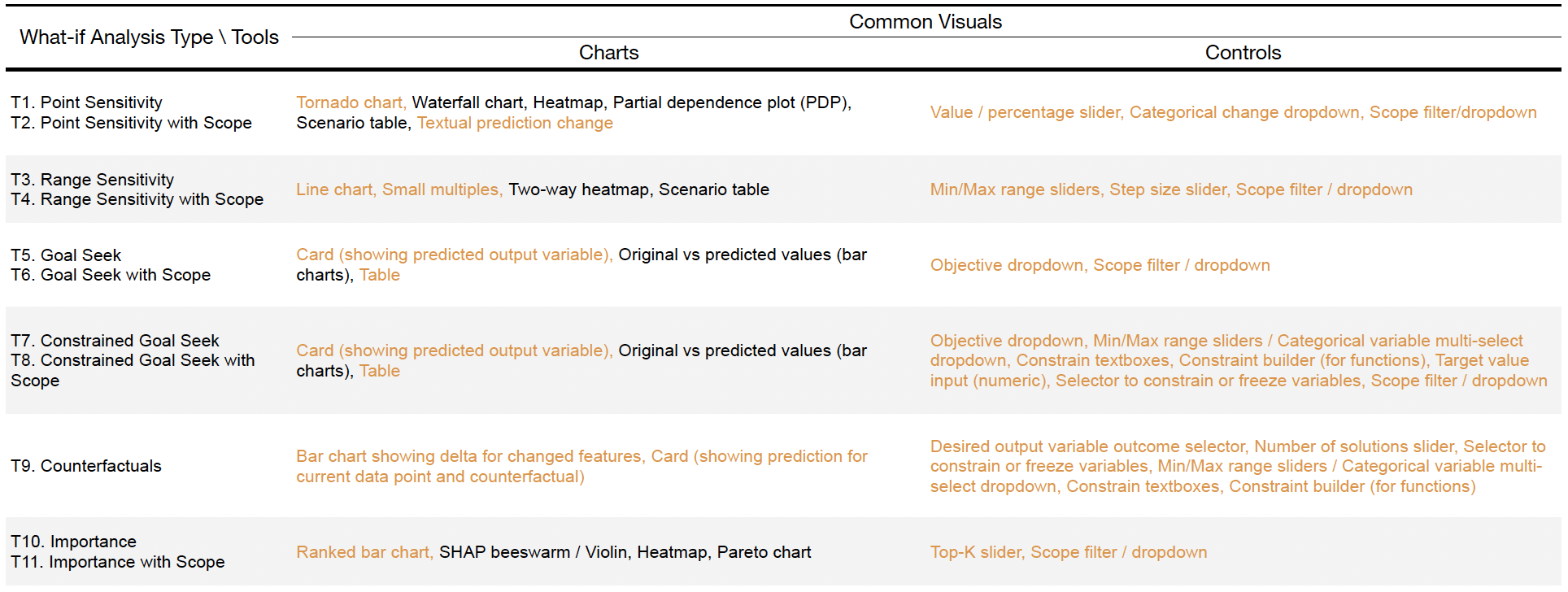}
    \caption{Common visualizations and controls across the 11 WIA types (T1--T11), observed in existing BI tools and research systems. The subset implemented by our compiler is highlighted; remaining entries are alternatives from prior work.}
    \label{fig:designspace}
    \Description{A table mapping each of the eleven WIA types to typical visual encodings and interaction controls. Columns list visualizations (such as bar charts, line charts, tornado charts, prediction cards, ranked bars, and heatmaps) and controls (such as value and range sliders, dropdowns, scope filters, and constraint builders). Entries implemented by our compiler are highlighted in orange; the remaining entries are alternatives observed in prior BI tools and research systems.}
\end{figure*}

Our compiler translates a \psl specification deterministically into an interactive interface through three stages: parsing, execution, and rendering.
First, the specification is \textit{parsed} to extract primitives, validate schema constraints, and populate missing defaults; structural violations are caught here and reported against the specific offending property, before any computation runs.
Second, type-specific analysis functions are \textit{executed} against the model and data as the specification dictates.
Crucially, the \codeN{scope} property filters the data \textit{before} the model is trained and applied, so a scoped analysis is genuinely restricted to the intended subset--not merely used to set a baseline--and the \codeN{constraints} and \codeN{objective} properties drive the goal-seek search directly, so optimization targets exactly the criterion the specification names.
Because these behaviors are fixed in the compiler rather than re-derived by an LLM for each question, they are consistent across questions and runs--the failure modes of \autoref{sec:direct_NL_interfaces}, where scope silently failed to shape the model and goal-seek searches were degenerate, cannot arise by construction.
Third, results are \textit{rendered} using deterministic visual mappings: each WIA type maps to a fixed visual encoding and set of controls, drawn from a design space of visualizations and controls observed across existing BI tools and research systems~\cite{gathani2025if,wexler2019if,hohman2019gamut,gathani2026praxa} (\autoref{fig:designspace}).
For each type we use typical visual encodings (e.g., bar charts for point sensitivity, line charts for range sensitivity, prediction cards for goal-seek, ranked bars for importance) with corresponding interaction controls (sliders for continuous variables, dropdowns for categorical variables, filters for scope, and constraint builders for goal-seek); our compiler implements the highlighted subset, while the remaining entries indicate alternatives available in prior systems.
User interactions update the underlying specification and trigger re-execution, keeping the interface and the analysis logic in sync.
Because this property-to-component mapping is deterministic, the same specification always yields the same interface, every interface element is traceable to the specification property that produced it, and this same mapping defines the surface through which specification errors become visible in the compiled interface--the basis for the error-propagation analysis in \autoref{sec:intermediate_NL_interfaces}.

\section{Evaluating \psl: Generation, Repair, and Error Propagation}
\label{sec:intermediate_NL_interfaces}
The motivating experiment (\autoref{sec:direct_NL_interfaces}) showed that direct generation fails precisely because analytical intent has no explicit representation: misread objectives, substituted scopes, and dropped constraints became embedded in code where they could not be seen, checked, or fixed.
We now evaluate whether routing generation through \psl--which makes that intent explicit--changes this.
We ask three questions: how reliably LLMs translate NL WIA questions into \psl specifications and where they fail (\textbf{RQ1}); whether the explicit structure of \psl enables errors to be detected and repaired (\textbf{RQ2}); and how residual errors propagate into compiled interfaces (\textbf{RQ3}).
All three use the benchmark and six-component ground-truth intent from \autoref{sec:benchmark}, encoded as \psl for this evaluation (\autoref{sec:groundtruth}).

Two points connect this evaluation to the motivating experiment and frame everything that follows.
First, the failures are largely the \textit{same}: the misread objectives, substituted scopes, and dropped constraints that made directly generated interfaces wrong reappear here as errors in \psl generation, confirming that faithfully translating a WIA question is hard regardless of the target.
What changes is not whether these errors occur but what can be done about them---in the motivating experiment, diagnosing why an interface was wrong required reading generated React and Python line by line, whereas here the same misinterpretations are visible \textit{in the specification itself}, as wrong values of named properties.
Second, whereas the motivating experiment sampled 150 questions per model to permit interface-level hand-scoring, this evaluation covers the full benchmark ($405$ questions $\times$ 3 models $=1{,}215$ specifications), auditing each against ground-truth intent at the property level.
The two studies are therefore complementary rather than comparable: one measures how often direct generation fails and how those failures hide; the other measures, at scale, whether an explicit representation makes the same class of failures tractable.

\subsection{RQ1: Generating \psl Specifications}
\label{subsec:generation}
\begin{figure*}[t]
    \centering
    \includegraphics[width=\textwidth]{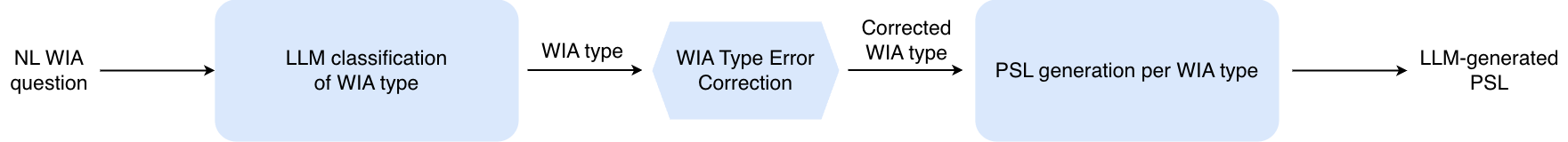}
    \caption{Strategy adopted for generating \psl for NL WIA questions in benchmark.}
    \label{fig:6_automation_strategy}
    \Description{A left-to-right flow diagram of the PSL generation strategy. A natural-language WIA question passes to an LLM classification step that assigns a WIA type, followed by a type-error correction step that yields a corrected type, followed by PSL generation conditioned on that type, producing the final LLM-generated PSL specification.}
\end{figure*}

We evaluate three LLMs (GPT-4o, GPT-5, and Claude-Sonnet-4) via a two-step few-shot prompting strategy (\autoref{fig:6_automation_strategy}).

\subsubsection{Specification Generation} 

\bpstart{Step 1: Classification of WIA Type}
Each LLM classified questions into one of 11 WIA types using prompts with dataset context, type definitions, and a few examples.
To reduce bias, we randomized type order, collected three predictions per question, and used majority vote.
Three coders annotated types as a human baseline.

\textbf{\textit{Findings.}}
Across 405 questions, we observed 119 LLM--human mismatches (details in supplementary material).
Most (102/119) were LLM misclassifications, which typically over-interpreted phrasing or failed to distinguish scoped from unscoped analyses.
In the remaining 17 cases, LLMs were more accurate than humans (e.g., correctly classifying a continuous range question as range sensitivity (T3) rather than sensitivity (T1)).
We reviewed all 119 cases and adopted the more accurate label before specification generation, so that the generation errors analyzed below are not confounded by upstream type-classification disagreements.

\bpstart{Step 2: \psl Specification Generation}
Using the predicted WIA type, each LLM generated a \psl specification via prompts containing dataset context, the assigned type, its definition, the \psl schema, and two to four hand-authored question--specification pairs.\footnote{Exemplars were fixed in advance and shared across all models and WIA types; we report generation reliability against this frozen exemplar set and do not tune prompts per model.}
Two coders compared each generated specification against the ground-truth intent at the property level, considering dataset and question context.
This process also surfaced schema gaps (e.g., expressive scoping: ``For customers with higher estimated salary\ldots'' required support for complex functions within the \codeN{scope} property, such as \codeN{"EstimatedSalary": \{"operator":} \codeO{">="}\codeN{, "function":} \codeO{"quartile3"}\codeN{\}}), leading to extensions of the \codeN{scope} property and re-generation.

\textbf{\textit{Findings.}}
Across all $1{,}215$ model-generated specifications, 637 (52.42\%) were generated correctly without intervention, matching the ground truth; 578 were erroneous prior to intervention (\autoref{fig:7_errors}; per-dataset breakdown in supplementary material).
That roughly half of specifications are wrong on the first pass--even with an explicit target schema--reinforces the lesson of the motivating experiment: the hard part of NL-driven WIA is not rendering an interface but faithfully capturing intent, and an explicit representation does not make LLMs better analysts.
Its value, as the rest of this section shows, is that it changes what happens to the errors that remain: instead of being entangled in code, they surface as wrong values of specific, named properties.

\begin{figure}[t]
    \centering
    \includegraphics[width=\columnwidth]{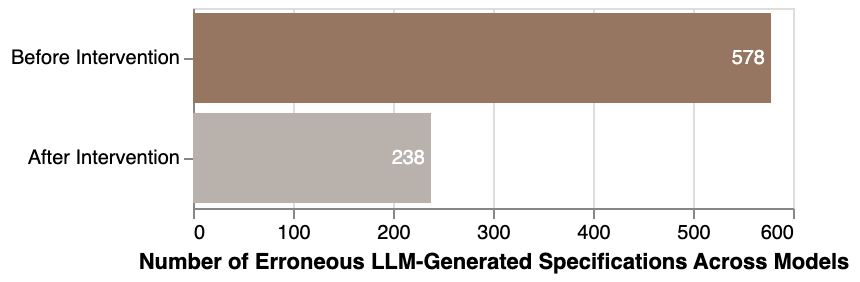}
     \caption{Number of erroneous LLM-generated specifications across models before and after intervention of targeted repair.}
    \label{fig:7_errors}
    \Description{A horizontal bar chart comparing the total number of erroneous LLM-generated specifications before and after targeted repair, aggregated across the three models. The "before intervention" bar reaches 578 erroneous specifications and the "after intervention" bar drops to 238, showing a substantial reduction after repair.}
\end{figure}

\subsubsection{Error Taxonomy}
\begin{figure*}[t]
    \centering
    \includegraphics[width=\textwidth]{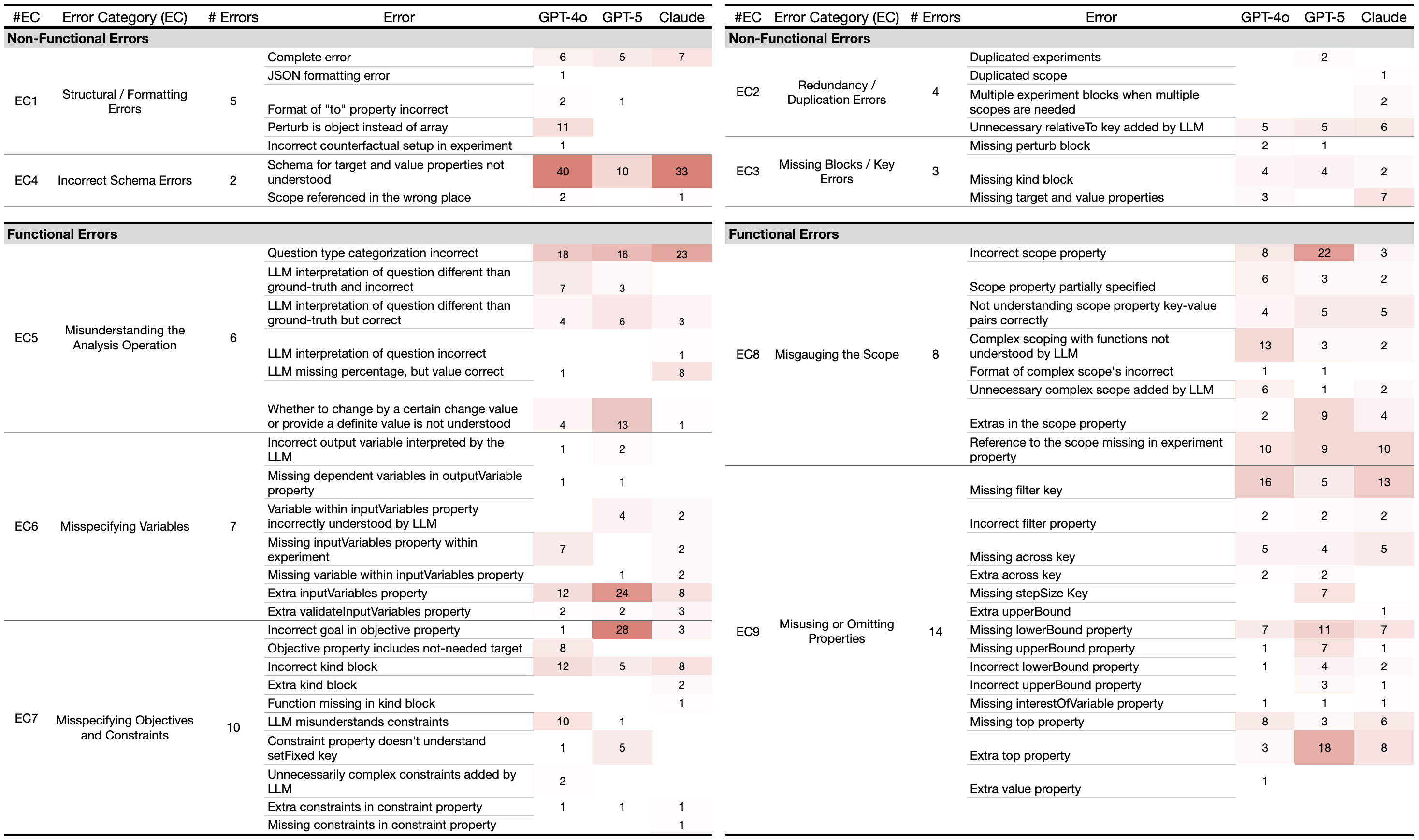}
    \caption{We identify 2 classes of errors observed in the LLM-generated specifications: (1) \textit{non-functional errors} where the specifications cannot be parsed and (2) \textit{functional errors} where specifications can be parsed but may not express the correct or entire intent of the what-if questions. For each of these classes, we show a number of error categories (EC), its selected examples of errors within them, and percentage of total specifications showing the error across the three LLMs.}
    \label{fig:8_errors_before_intervention}
    \Description{A two-part table of the error taxonomy. The left region lists non-functional error categories EC1 to EC4 (structural or formatting errors, redundancy or duplication, missing blocks or keys, and incorrect schema errors); the right region lists functional error categories EC5 to EC9 (misunderstanding the analysis operation, misspecifying variables, misspecifying objectives and constraints, misgauging scope, and misusing or omitting properties). Each category shows selected example errors and the percentage of specifications exhibiting them for GPT-4o, GPT-5, and Claude, with functional errors far more common than non-functional ones.}
\end{figure*}

To analyze failures, we audited the erroneous specifications across all datasets and models in two passes.
At the \textit{parsing level}, we checked schema validity and compilability; failures here are \textit{non-functional errors}, since they cannot produce any output.
At the \textit{semantic level}, we evaluated whether parsed specifications captured the intended analysis; failures here are \textit{functional errors}, since they compile but produce incorrect outputs.
We organize the errors observed \textit{at generation} into nine categories (EC1--EC9) across these two classes; a single specification may exhibit multiple errors (\autoref{fig:8_errors_before_intervention}).
A tenth category, introduced only by repair, is described in \autoref{subsec:repair}.

\bpstart{Non-Functional Errors (EC1--EC4)}
Of the 578 incorrect specifications, 58 failed to compile into executable interfaces, falling into four categories.

\textit{EC1: Structural / Formatting Errors.}
These are the classic ``won't run'' failures: the LLM not returning any output at all (labeled as complete error in the figure), malformed JSON, dangling commas or quotes, and container-shape mistakes such as emitting a single \codeN{perturb} object where an array is required.
These appear infrequently and are shallow syntax issues; a smaller fraction are structural mismatches where the structure of a specific WIA schema is incorrect (e.g., a counterfactual missing the feature name to which it wants the closest data point).

\textit{EC2: Redundancy / Duplication Errors.}
The model sometimes generates redundant content--e.g., duplicated \codeN{experiment} blocks, near-identical blocks regenerated, or hallucinated non-schema keys (e.g., \codeN{relativeTo}) that the model invents.
These errors are infrequent but costly, creating confusion over which block to use, inflating computation, or producing inconsistent results.
We also observed cases where redundant blocks masked missing or incorrect parts of the specification, complicating debugging.

\textit{EC3: Missing Blocks / Key Errors.}
Here, although the generated specification is well-formed JSON, it omits a critical block the schema requires--for example, missing \codeN{target}, \codeN{value}, or \codeN{kind} inside a goal-seek experiment, or the entire \codeN{perturb} block for a sensitivity experiment.
These errors are rare, yet they behave like EC1 at runtime, where the pipeline aborts.

\textit{EC4: Incorrect Schema Errors.}
This is the most prominent non-functional category, referring to violations of the schema during generation.
Most notable was swapping the \codeN{target} and \codeN{value} roles, or referencing \codeN{scope} from unsupported positions in the \codeN{experiment} block.
These mistakes are subtle since on the surface the JSON is clean, but schema validation fails or the engine refuses to run because required invariants (e.g., \codeN{kind} needs \codeN{target} and \codeN{value} performing certain roles) are not satisfied.

\bpstart{Functional Errors (EC5--EC9)}
Unlike non-functional errors, these specifications compile and produce plausible interfaces but are incorrect because they misinterpret the question (e.g., wrong data subset or constraints).
They account for the majority of failures (520 of 578), and they are the direct analogue of the silent failures in \autoref{sec:direct_NL_interfaces}--the same misread objectives, scopes, and constraints, now surfaced as wrong properties rather than buried in code.
We group them into five categories (\autoref{fig:8_errors_before_intervention}).

\textit{EC5: Misunderstanding the Analysis Operation.}
This error captures cases where the specification is structurally valid but semantically mismatches the intended scenario, due to incorrect mapping from NL intent to specification semantics.
Common failures include paraphrasing the question, especially in compound cases with implicit constraints (e.g., ``within budget''), multiple perturbations (e.g., ``increase A and decrease B but hold C fixed''), or ambiguous quantifiers (e.g., ``small increase'').
LLMs may misinterpret these, leading to deviations from ground truth.
Additional errors include misreading perturbations (e.g., relative vs.\ absolute changes, percentage vs.\ value) and omitting required experiments.
Even when the analysis type is pre-specified correctly, LLMs occasionally generate incorrect \psl (e.g., treating a constrained goal-seek as a sensitivity analysis).

\textit{EC6: Misspecifying Variables.}
The generated \psl runs but uses the wrong features in the dataset or omits the right ones.
These include an incorrect \codeN{outputVariable} chosen for the experiment, missing \codeN{outputVariables} when the question inquires about more than one feature, a model choice inappropriate for the specified \codeN{outputVariable}, forgetting to include the \codeN{inputVariables} the question inquires about, restricting focus to only some \codeN{inputVariables}, missing or misinterpreting variable names within \codeN{inputVariables}, confusing ``change by'' vs.\ ``set to'' semantics within \codeN{perturb}, or spurious properties unknown to the schema (e.g., \codeN{validateInputVariables}).

\textit{EC7: Misspecifying Objectives and Constraints.}
These errors arise in goal-seeking and constrained experiments, where the specification must encode both a desired outcome and the bounds within which the system should search.
Most involve misspecified objectives--minimizing when the question asks to maximize, adding optimization targets where none are needed, or failing to express complex target functions (e.g., ``solutions where Revenue increases by 10\%'').
Incorrect or duplicate \codeN{kind} blocks cause the optimization to converge to the wrong criterion, producing an interface that displays an ``optimal'' solution optimizing the wrong thing.
Other errors involve constraints: reversed inequality bounds (e.g., $\geq$ instead of $\leq$), constraints applied to the wrong variable (e.g., bounding \textit{Revenue} when the question constrains \textit{Spend}), misuse of the \codeN{setFixed} property, and gratuitous complexity where the LLM introduces redundant nested constraints not present in the question.
The resulting interface enforces incorrect feasibility bounds, producing recommendations that violate the analyst's intended restrictions or exclude valid solutions.

\textit{EC8: Misgauging the Scope.}
These errors arise when analyses are restricted to data subsets but the \codeN{scope} is malformed, incomplete, or extraneous, altering which rows are evaluated.
Common issues include incorrect key--value mappings, especially for categorical features encoded numerically (e.g., \textit{Email\_Type} as 1 or 2 vs.\ ``Transactional'' or ``Promotional'').
LLMs also generate unsupported or over-engineered expressions (e.g., SQL-like clauses, regex, or unnecessary nested logic).
Another recurring issue is that the \codeN{experiment} block fails to reference the defined scope, causing filters to be ignored.
As a result, analyses run on incorrect subsets, yielding misleading conclusions.

\textit{EC9: Misusing or Omitting Properties.}
This is one of the most pervasive errors.
The generated specification often has missing properties (e.g., \codeN{stepSize}, \codeN{lowerBound} and \codeN{upperBound} in sensitivity experiments, or \codeN{top} for importance experiments) or incorrect property values (e.g., not within the dataset bounds), which distort the search space of experiments.
Conversely, additional unwarranted properties can yield plausible but misleading results.

\textbf{\textit{The taxonomy is itself the evidence for inspectability.}}
Every one of these categories is tied to a \textit{named property} of the specification, and this is not incidental---it is the central advantage of an explicit representation made concrete.
The entire audit above, assigning each of 578 failures to specific components, was possible only because intent is readable in the specification: the same audit on the directly generated interfaces of \autoref{sec:direct_NL_interfaces} would have required reading hundreds of bespoke programs line by line, with no guarantee of finding where--or whether--intent was misencoded.
Here it reduces to inspecting a handful of named fields against the question.
That same localizability is what makes the errors not only detectable but repairable, which we turn to next.

\subsection{RQ2: Detecting and Repairing Errors}
\label{subsec:repair}
Having shown that errors localize to named properties, we now ask whether they can be found automatically and fixed.
We find that identification is best left to humans and repair to the model---a division of labor the explicit representation uniquely enables.

\begin{figure}[t]
    \centering
    \includegraphics[width=\columnwidth]{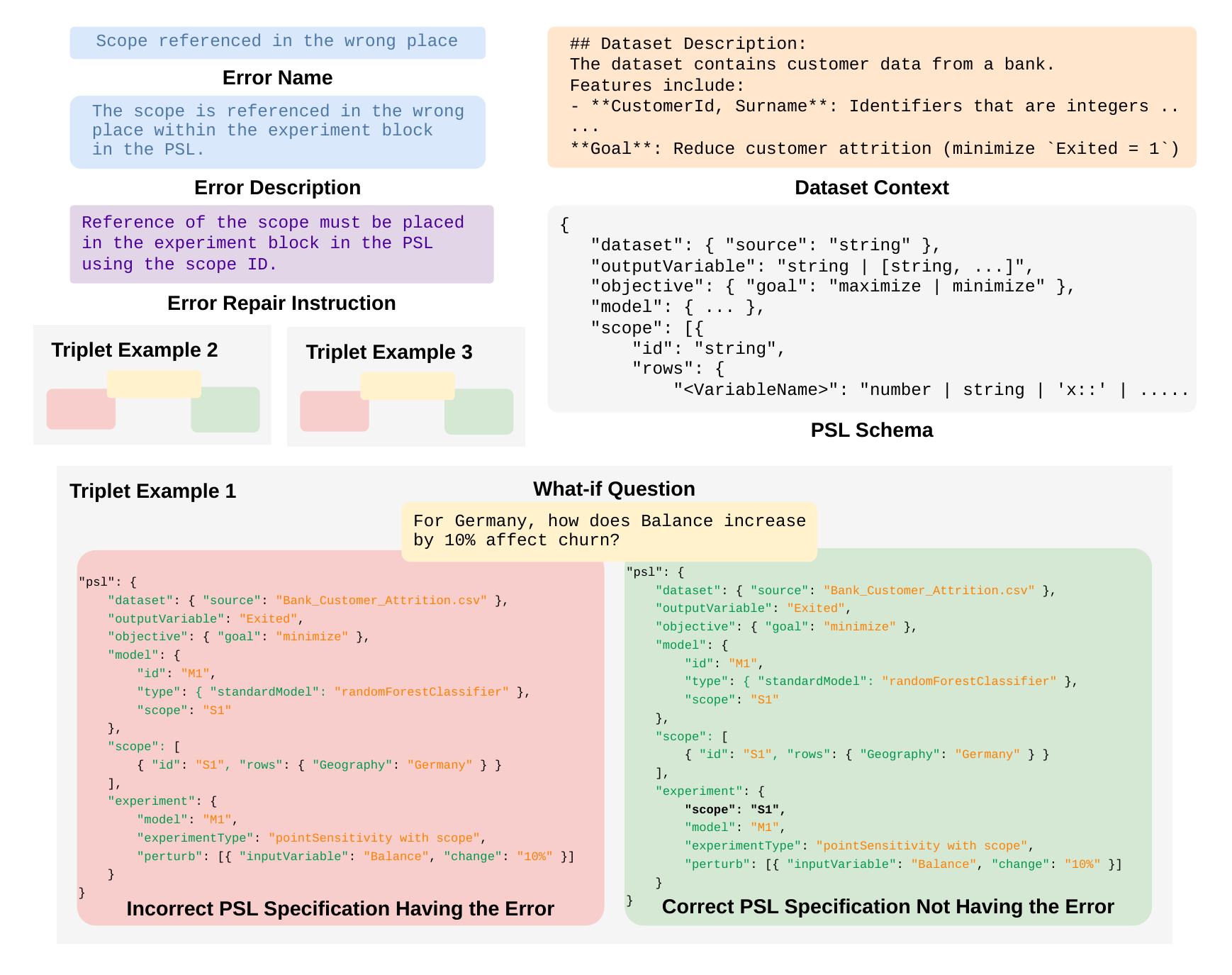}
    \caption{Example of a targeting prompt contents for correcting the ``Scope referenced in the wrong place'' error.}
    \label{fig:9_repair}
    \Description{An annotated example of a targeted repair prompt for the "scope referenced in the wrong place" error. It shows the error name and description, a short repair instruction, dataset context, the PSL schema, and paired triplets of a what-if question with an incorrect PSL specification (with scope referenced inside the experiment block) beside the corrected specification (with scope referenced correctly).}
\end{figure}

\begin{figure*}[t]
    \centering
    \includegraphics[width=\textwidth]{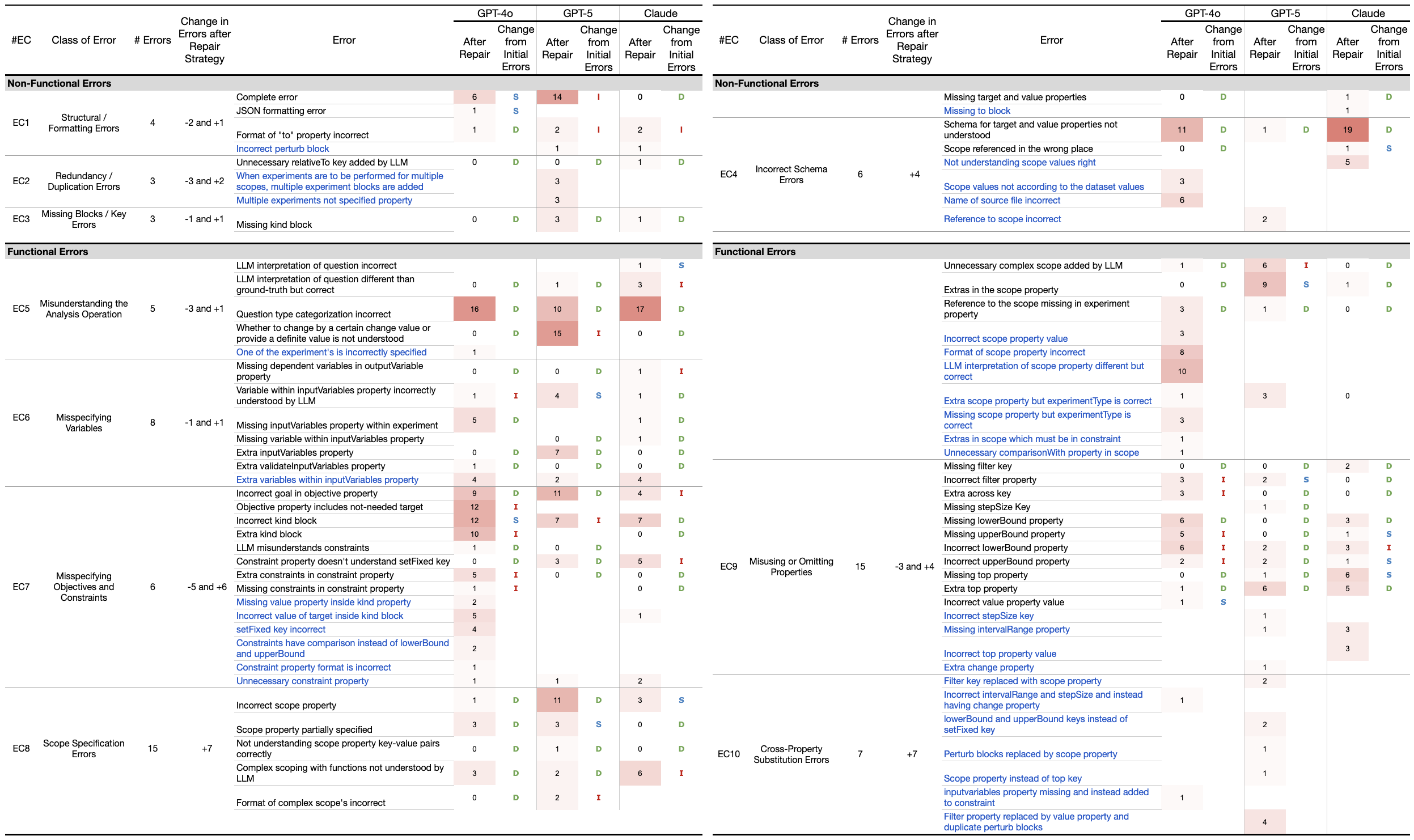}
    \caption{Error distribution after targeted repair across error categories (EC1–EC10) and models. For each error, we report the count after intervention and the direction of change from before intervention (D = decrease, I = increase, S = same). New errors introduced during repair are marked in blue. Most categories decrease, but new error types emerge like cross-property substitutions (EC10).}
    \label{fig:10_after_intervention}
    \Description{A detailed table of error counts after targeted repair, broken down by the ten error categories EC1 to EC10 and by model (GPT-4o, GPT-5, Claude). Each cell reports the post-repair count and a direction-of-change indicator (D for decrease, I for increase, S for same), with newly introduced errors marked in blue. Most categories decrease after repair, but scope and schema errors increase and a new category, cross-property substitutions (EC10), appears entirely from repair.}
\end{figure*}

\subsubsection{Automated Error Detection}
We evaluate whether LLMs can detect errors in generated specifications through two experiments.
In the first (\textit{binary detection}), we asked LLMs to predict whether a specification contained any error (yes/no), compared against majority-vote labels from three human annotators.
Across all 405 questions, LLMs achieved 64.06\% accuracy with modest agreement ($\kappa = 0.218$), but over-flagged errors (64.06\% vs.\ 44.43\% by humans).
This suggests LLMs can act as high-recall screeners that surface likely-problematic specifications for human review, but are not reliable as final decision-makers.

In the second experiment (\textit{per-category diagnosis}), we tested whether LLMs could identify the specific error category (EC1--EC9) when provided with error definitions and contrastive positive and negative examples alongside each specification.
On a sample of 420 specifications (140 questions $\times$ 3 models), overall agreement was only fair ($\kappa \approx 0.23$--0.25), with LLMs over-flagging at 3--23$\times$ the human rate.
LLMs were better calibrated on functional errors (EC5--EC9) than non-functional ones (EC1--EC4).
Full details are in supplementary material.

These findings show that LLMs can detect \textit{that} a specification is likely incorrect but struggle to diagnose \textit{what} is wrong, necessitating human input for precise error categorization.
This motivates a human-in-the-loop division of labor--a human identifies the error category, and the LLM applies a category-specific fix.
Crucially, that division of labor is only meaningful because the representation is inspectable: a human can identify \textit{which} named property is wrong by reading the specification against the question---an act that has no counterpart when intent is dispersed through generated code.

\subsubsection{Targeted Repair}
For each erroneous specification (N = 578 across 3 models), we prompted the generating LLM with a tailored repair prompt including: (1) the error name and a one-sentence description, (2) a short error-specific repair instruction, (3) 2--3 triplets of question, incorrect specification, and corrected specification demonstrating the fix, (4) dataset context, and (5) the \psl schema (\autoref{fig:9_repair}).
Two coders then re-evaluated the repaired specifications against ground truth.

\textbf{\textit{Findings.}}
Repair reduced erroneous specifications from 578 to 238, raising overall correctness from 52.42\% (before intervention) to 80.42\% (after intervention) (\autoref{fig:7_errors}).
The significance is less the specific gain than what makes it achievable: each repair targets a \textit{named} property identified as wrong--a missing \codeN{scope}, an inverted \codeN{objective}, an out-of-range \codeN{top}--rather than an unlocalized defect somewhere in generated code.
The same errors, when embedded in the directly generated interfaces of \autoref{sec:direct_NL_interfaces}, offered no comparable point of intervention: there is no property to correct, only code to rewrite, and no way to confirm the rewrite preserved everything else.

At the category level, we observed two opposing effects (\autoref{fig:10_after_intervention}).
First, several previously observed errors decreased or were eliminated.
Among non-functional errors, structural formatting issues such as complete failures and malformed JSON decreased across all models (EC1, $-2$ net), and missing blocks were resolved (EC3, $-1$ net).
Among functional errors, question-type miscategorization and value-vs-change confusion decreased (EC5, $-3$ net), several variable-specification errors were resolved including missing \codeN{inputVariables} and incorrect \codeN{outputVariable} interpretations (EC6, $-1$ net), and property omissions such as missing \codeN{lowerBound} and \codeN{top} were partially addressed (EC9, $-3$ net).

Second, new errors emerged.
Incorrect schema errors increased (EC4), driven by new failures in understanding scope values and dataset-specific encodings.
Scope specification errors grew substantially (EC8), with new patterns including incorrect scope property values, misunderstood scope formats, and LLM-interpreted scope properties differing from ground truth.
Objective and constraint errors also increased (EC7, $-5$ resolved but $+6$ new), with new errors including incorrect \codeN{kind} block values, \codeN{setFixed} misuse, and constraint format issues.
Most notably, a new category emerged that was absent at generation and introduced only by repair--\textit{cross-property substitutions} (EC10, $+7$ entirely new)--where LLMs replaced one structural construct with another (e.g., \codeN{perturb} blocks replaced by \codeN{scope}, \codeN{filter} replaced by \codeN{value} while duplicating \codeN{perturb}, or \codeN{inputVariables} removed and replaced with constraints).
These patterns suggest that when LLMs repair a targeted property, they inadvertently edit adjacent properties, introducing higher-order faults.

These results highlight both the strength and the current limits of the intermediate representation for repair.
The explicit structure of \psl makes first-order errors localizable and fixable through targeted prompts.
Preventing higher-order cross-property drift, however, likely requires stronger guardrails--block-level schema validation after each repair, slot-filling approaches that modify only the targeted property rather than regenerating the full specification, or richer contrastive exemplars with explicit instructions.
Notably, even these mitigations are only expressible \textit{because} the representation is explicit: each names a property or block to constrain---a guardrail that could not be stated, let alone enforced, over free-form generated code.

\subsection{RQ3: How Errors Propagate into Interfaces}
\label{sec:rq3}
\begin{figure*}[t]
    \centering
    \includegraphics[width=\textwidth]{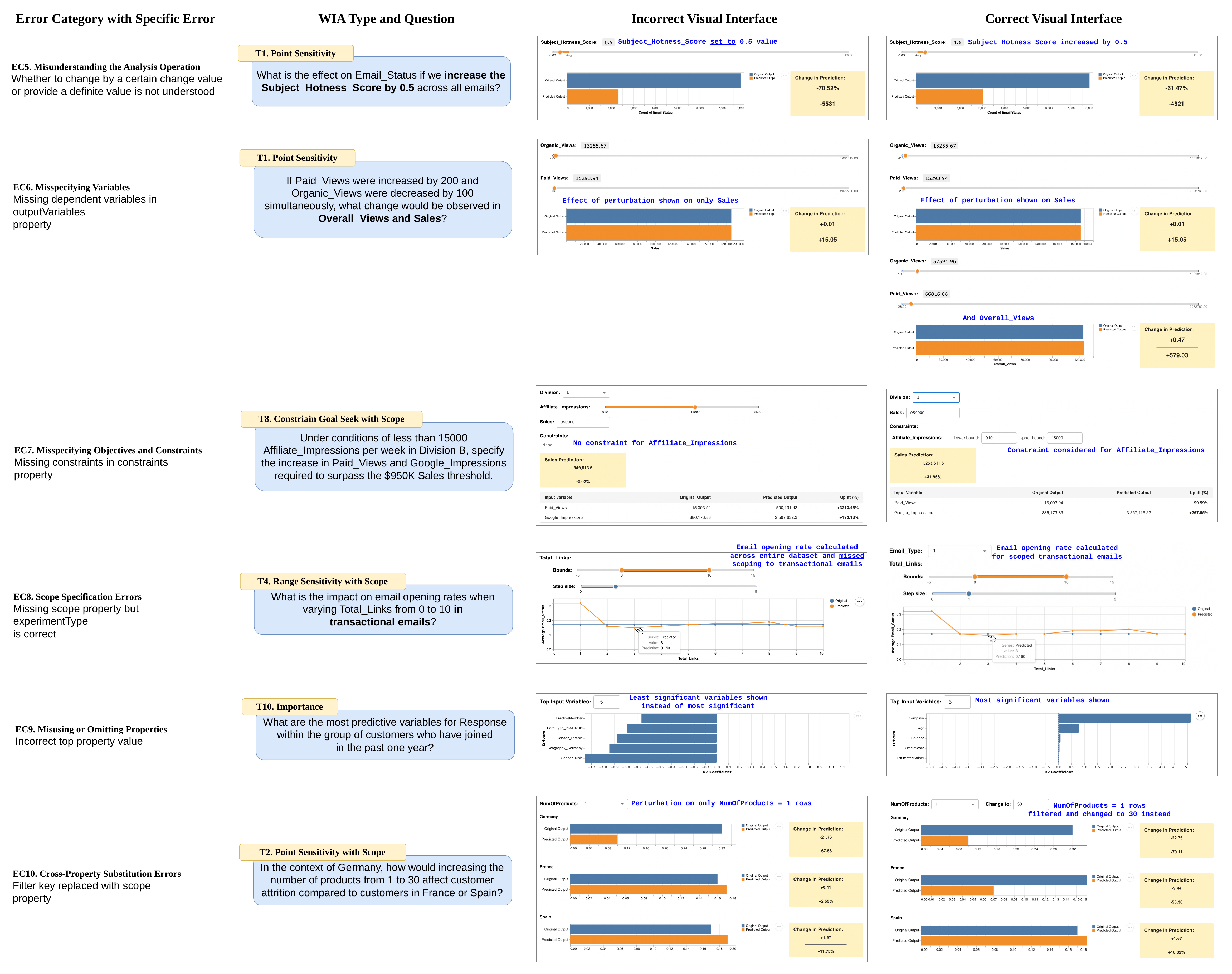}
    \caption{Example of an error across all functional errors (EC5–EC10) and their impact on the visual interface. We show interfaces produced by erroneous specifications (incorrect behavior) and instead the intended corrected interfaces if specifications are repaired.}
    \label{fig:11_interfaces}
    \Description{A multi-row figure with one row per functional error category EC5 to EC10. Each row shows the what-if question, the specific specification error, a screenshot of the incorrect interface produced by the erroneous specification, and a screenshot of the corrected interface from a repaired specification. Examples include a slider set to an absolute value instead of a relative change, a missing output variable, a dropped constraint, a missing scope filter, an inverted top value showing least instead of most important variables, and a filter replaced by a scope property. In each case the incorrect and correct interfaces look visually similar despite encoding different analyses.}
\end{figure*}

The repair analysis shows that most errors can be caught and fixed; RQ3 examines what happens to the ones that are not.
These residual functional errors--specifications that compile and render but misrepresent intent--are the direct counterpart of the silent failures in \autoref{sec:direct_NL_interfaces}: interfaces that look right but are wrong.
The difference is that here each has a traceable cause.
As described in \autoref{sec:workflow}, compilation is deterministic, so any error in a specification propagates directly into the interface and every interface defect is traceable back to the property that produced it.
\autoref{fig:11_interfaces} shows, for each functional error category, the WIA question, the specific specification error, the incorrect interface produced, and the correct interface from a repaired specification.

\bpstart{Change-by vs.\ set-to (EC5)}
The question asks to increase \texttt{Subject\_Hotness\_Score} \textit{by} 0.5, but the specification sets the value \textit{to} 0.5.
The incorrect interface shows a slider set to 0.5 (absolute), while the correct interface increases the baseline by 0.5--yielding different predictions that mislead the analysis.

\bpstart{Missing a dependent variable (EC6)}
Asked about the effect of changing \texttt{Paid\_Views} (by $+200$) and \texttt{Organic\_Views} (by $-100$) on \textit{both} \texttt{Overall\_Views} and \texttt{Sales}, the specification omits \texttt{Overall\_Views} from \codeN{outputVariables}.
The interface shows the effect on \texttt{Sales} only--a partial answer, with nothing signalling the omission.

\bpstart{Missing constraints in constrained goal-seek (EC7)}
For a constrained goal-seek requiring ``less than 15{,}000 Affiliate\_Impressions per week,'' the specification omits the constraint entirely, while the correct interface respects it and highlights it visually for transparency.

\bpstart{Missing scope (EC8)}
The question targets ``transactional emails,'' but the specification omits the \codeN{scope} property.
The incorrect interface computes email opening rates over the full dataset, whereas the correct interface filters to transactional emails only.

\bpstart{Inverted property value (EC9)}
Asked for the \textit{most} predictive variables, the specification sets \codeN{top} to $-5$ instead of $+5$, so the interface surfaces the \textit{least} significant variables--rendered identically to a correct ranking.

\bpstart{Cross-property substitution (EC10)}
The question asks to increase number of products \textit{from} 1 to 30 while keeping all other rows, requiring a \codeN{filter} followed by \codeN{perturb}.
During repair, the LLM replaced the \codeN{filter} with the \codeN{scope} property, so the incorrect interface applies the perturbation only to rows with number of products equal to 1, rather than perturbing those rows and comparing across geography as intended.

These are exactly the failures of \autoref{sec:direct_NL_interfaces}--interfaces that render, respond, and look plausible while encoding a wrong analysis.
The decisive difference is that here the wrongness is not lost in code.
Because \psl makes analytical intent explicit as named properties, each of these defects traces to a specific component--a missing \codeN{scope}, an inverted \codeN{objective}, a substituted \codeN{filter}--and can be caught by inspecting the specification before the interface is ever rendered.
Across the three analyses, one thread holds: an explicit representation does not stop LLMs from misinterpreting WIA questions--the same misinterpretations recur throughout--but it converts those misinterpretations from silent, code-buried failures into named, inspectable, and repairable ones.

\section{Discussion}
\label{sec:discussion}
\bpstart{An explicit representation targets exactly where direct generation fails}
The central finding connecting our two studies is that the failures of direct generation and the strengths of an explicit representation concern the \textit{same} properties.
Direct generation most often mishandled objective and constraints (\autoref{sec:direct_NL_interfaces})--precisely the aspects of a question with no forced representation in generated code, so a model can omit them while the interface still renders plausibly.
\psl makes these same aspects named, first-class properties, and it is exactly these that targeted repair is able to localize and fix, raising correctness from 52.42\% to 80.42\% (\autoref{subsec:repair}).
Where a conflated scope condition in direct generation produced an unrecoverable crash, \psl's \codeN{scope} primitive is enforced structurally through compilation, removing that failure class by construction rather than by chance.
The properties hardest to get right without an explicit representation are thus the ones an explicit representation is best positioned to name, check, and repair.

\bpstart{Declarative specifications as intermediate representations for WIA}
More broadly, declarative specifications offer an effective middle ground between the manual effort of spreadsheet and BI tools and the opacity of direct code generation. By encoding analytical intent as named properties (e.g., \codeN{experimentType}, \codeN{objective}, \codeN{scope}, \codeN{perturb}, \codeN{constraints}), \psl represents intent in a form that is both executable and inspectable, extending to WIA the role SQL plays for queries and Vega-Lite for visualizations~\cite{satyanarayan2016vega,kim2022cicero}. 
Two properties make this valuable. First, intent is preserved as a persistent, named artifact rather than being entangled in implementation code, so it can be examined against the original question before any interface is produced.
Second, because compilation is deterministic, each property maps to a specific interface element, so intended behavior can be inspected prior to execution--a transparency absent from both manual formula construction and opaque code generation.

\bpstart{Human--AI collaboration, not full automation}
Our results point to a division of labor rather than end-to-end automation. LLMs generate correct specifications only about half the time (52.42\%) and, while they can flag \textit{that} a specification is likely wrong (64.06\% binary-detection accuracy), they cannot reliably diagnose \textit{which} property is at fault ($\kappa \approx 0.23$; \autoref{subsec:repair}).
Yet once a human identifies the error category, LLM-applied targeted repair raises correctness to 80.42\%.
This suggests LLMs are best used as high-recall generators and screeners, with humans providing high-precision categorization and verification.
Concretely, NL-driven WIA systems should expose the \psl specification alongside generated outputs, surface component-level signals that highlight likely errors, and support inline editing with immediate re-compilation--focusing human effort on high-impact judgments such as resolving ambiguity, validating constraints, and refining scope.

\bpstart{Why an explicit representation rather than an agentic loop}
An alternative to an explicit representation is an agentic generation loop that iteratively executes and self-corrects generated code. Such loops can remove incidental implementation faults, but they do not address the failures we observe: a dropped constraint or substituted objective produces code that runs and renders correctly, so there is no execution signal for the loop to correct against.
Our contribution is orthogonal to the generation method--the value of \psl is that it gives analytical intent an existence independent of the implementation, making these silent errors inspectable and repairable however the specification was produced.
An agentic loop and an explicit representation are therefore complementary rather than competing.

\bpstart{Limitations and future directions}
Several limitations bound our claims. 
First, our motivating experiment (\autoref{sec:direct_NL_interfaces}) scores semantic correctness with an automated comparison validated against human judgment on 40 interfaces ($\kappa = 0.85$); while agreement is strong, a larger human-scored evaluation would tighten the estimate, and we read the objective rates--where agreement was lowest--with corresponding caution. 
It also evaluates a stratified 150-question subsample per model rather than the full benchmark, a deliberate trade-off to permit manual verification. 
Second, and most importantly, our evaluation establishes that \psl makes intent inspectable and repairable at the \textit{system} level--errors are localizable to named properties and fixable through targeted repair--but does not establish whether \textit{end users} can read, validate, and repair \psl specifications in practice.
Whether \psl-grounded interfaces improve user understanding, task performance, and error recovery relative to direct generation is an open question we regard as the essential next step.
Third, cross-property substitution errors (EC10) show that targeted repair can introduce higher-order faults; lightweight inter-block dependency rules (e.g., ``\codeN{scope} must not wrap \codeN{filter}'') and slot-filling generation that edits only the targeted property~\cite{moritz2018formalizing} are promising mitigations.
Fourth, our benchmark covers 11 WIA types across 5 tabular datasets but excludes temporal, causal, and multi-step chained analyses~\cite{wongsuphasawat2017voyager}; extending \psl with temporal operators, causal structures, and nested experiments would broaden coverage while preserving the separation of what is analyzed from how it is implemented.
Although \psl is our instantiation, the two-stage workflow (NL-->structured representation-->interface) should generalize to other analytical tasks with well-defined structure~\cite{satyanarayan2014lyra,schulz2013design}.
Finally, our two evaluations use different Claude snapshots within the same model family--\texttt{claude-sonnet-4} for the workflow evaluation and \texttt{claude-sonnet-4.5} for the formative experiment--as provider snapshots are periodically retired and superseded. As with all LLM evaluations tied to dated snapshots, exact reproduction depends on provider availability; because our two evaluations are reported separately rather than compared, this difference does not affect our conclusions.
\section{Related Work}
\label{sec:related_work}
Our work relates to WIA systems, NL data analysis interfaces, and declarative intermediate representations.

\bpstart{What-if Analysis Systems}
WIA has been studied as a tool for exploring hypothetical scenarios and supporting data-driven decision-making.
Prior systems support WIA across domains, including business analytics~\cite{gathani2021augmenting,gathani2025if}, healthcare~\cite{bhattacharya2023directive,laguna2023explimeable}, social media~\cite{wu2014opinionflow}, environmental modeling~\cite{luo2017impact,hazarika2023haiva}, and many others.
These systems enable users to vary parameters, explore outcomes, and compare scenarios, but typically require substantial manual effort in implementation.
Recent work synthesizes prior WIA systems into a unified grammar of workflows through compositional primitives~\cite{gathani2026praxa}, and introduces \psl as a declarative encoding of this grammar.
That work establishes the grammar and illustrates its use as an intermediate representation, but does not evaluate whether it can reliably mediate between NL questions and interactive interfaces---whether LLMs can generate faithful specifications, what errors arise, or whether those errors can be repaired.
Our work operationalizes \psl in an end-to-end workflow--including LLM-based generation from NL, a taxonomy-driven error analysis, and targeted repair--and, prior to that, examines through a formative experiment whether an explicit intermediate representation is needed at all, rather than assuming it.

\bpstart{NL Interfaces for Data Analysis}
NL interfaces aim to lower the barrier to data analysis by allowing users to express intent in natural language~\cite{srinivasan2017natural}.
Systems such as Eviza~\cite{setlur2016eviza}, NL4DV~\cite{narechania2020nl4dv}, and Leva~\cite{zhao2024leva} support interactive visualization and multi-step analysis through conversational interaction.
Other systems focus on statistical analysis and data transformations, including Iris~\cite{fast2018iris}, Analyza~\cite{dhamdhere2017analyza}, DataTone~\cite{gao2015datatone}, and FlowSense~\cite{yu2019flowsense}.
More recent LLM-based tools generate charts and transformations from NL, including ChartGPT~\cite{tian2024chartgpt} and Data Formulator~2~\cite{wang2025data}, the latter observing that single-turn, fully-specified NL prompts are ill-suited to iterative authoring--a limitation an explicit, editable intermediate representation directly addresses.
Other applications include interface customizations~\cite{vaithilingam2024dynavis} and collaborative sensemaking~\cite{srinivasan2020inchorus}.

However, these approaches primarily support querying, visualization generation, or statistical operations, and do not capture the multi-step, model-driven nature of WIA.
They lack support for parameter manipulation, constraints, model-based reasoning, and iterative scenario exploration.
Our work addresses this gap by translating NL WIA questions into structured specifications that compile into interactive interfaces supporting these capabilities.

\bpstart{Declarative Specifications as Intermediate Representations}
Declarative representations separate what to compute from how to implement it, enabling transparency, validation, and systematic error handling across domains.
In databases, SQL serves as an interpretable intermediate representation for NL-to-SQL systems, supporting inspection and correction~\cite{tian2024sqlucid,tian2025text,fu2023catsql}.
In visualization, declarative grammars such as Vega-Lite~\cite{satyanarayan2016vega}, VizQL~\cite{hanrahan2006vizql}, and ggplot2~\cite{wickham2011ggplot2} enable systems to generate and manipulate visualizations through structured specifications~\cite{luo2021synthesizing,song2022rgvisnet,tian2024chartgpt}.
Declarative approaches have also been applied to interface generation and interactive documents~\cite{kin2012proton,heer2023living,chartifact}.
Building on prior work in declarative representations, we introduce \psl, a declarative intermediate representation grounded in the \praxa framework~\cite{gathani2026praxa}.
As shown in this paper, \psl captures model-driven, iterative WIA workflows and serves as a structured bridge between NL questions and interactive interfaces, preserving analytic intent, consistent with how analysts engage with declarative grammars in practice~\cite{py2023how}.
This explicit structure further enables targeted repair of LLM-generated specifications, extending declarative interaction design to NL-driven analytical workflows~\cite{satyanarayan2014declarative}.
\section{Conclusion}
We asked whether generating interactive WIA interfaces requires an explicit representation of analytical intent, or whether a capable LLM can produce them directly.
A motivating experiment across three state-of-the-art LLMs showed that direct one-shot generation fails both visibly--producing no working interface in 17--45\% of cases--and silently, rendering plausible interfaces that misrepresent the question's objective and constraints, with errors detectable only by reading the underlying code.
Motivated by these findings, we presented a two-stage workflow that first translates a WIA question into a \psl specification--making analytical intent explicit, inspectable, and independent of implementation--and then compiles it into an interactive interface.
On a benchmark of 405 questions across 11 WIA types and 5 datasets, LLMs generate correct specifications in 52.42\% of cases; because errors are localized to named specification properties, taxonomy-guided targeted repair raises this to 80.42\%, and we show how undetected errors would otherwise propagate into plausible but misleading interfaces. 
Together, these results show that the difficulty of reliable NL-driven WIA lies not in rendering interfaces but in faithfully capturing intent--and that an explicit intermediate representation is what makes that intent auditable and repairable.


\bibliographystyle{ACM-Reference-Format}
\bibliography{sample-base}

\newpage
\section{Supplementary}
\label{sec:supplementary}

\subsection{Formative Exploration Details}
\label{sec:supp_formative}
This section provides additional detail on the formative exploration of spreadsheet and BI tools reported in the main paper (Section 2): the dataset used, the full set of scenarios, and the procedure followed.
The four resulting challenges (C1--C4) are reported in full in Section 2.
(LLM-based generation is examined separately as a formative experiment in Section 4 and is not part of this exploration.)

\subsubsection{Dataset}
The \textit{Bank Customer Attrition} dataset~\cite{bankcustomerattritiondataset} was used for exploration.
Features in the dataset involve customer demographics (e.g., \textit{Age, Gender, Geography}), account-related features (e.g., \textit{CreditScore, Tenure, Balance, NumOfProducts, HasCrCard, IsActiveMember, EstimatedSalary}), service feedback metrics (e.g., \textit{Complain, Satisfaction Score, Card Type, Points Earned}), and a binary churn outcome (\textit{Exited}).

\subsubsection{WIA Scenarios Used in the Formative Exploration}
\begin{figure}[t]
    \centering
    \includegraphics[width=\columnwidth]{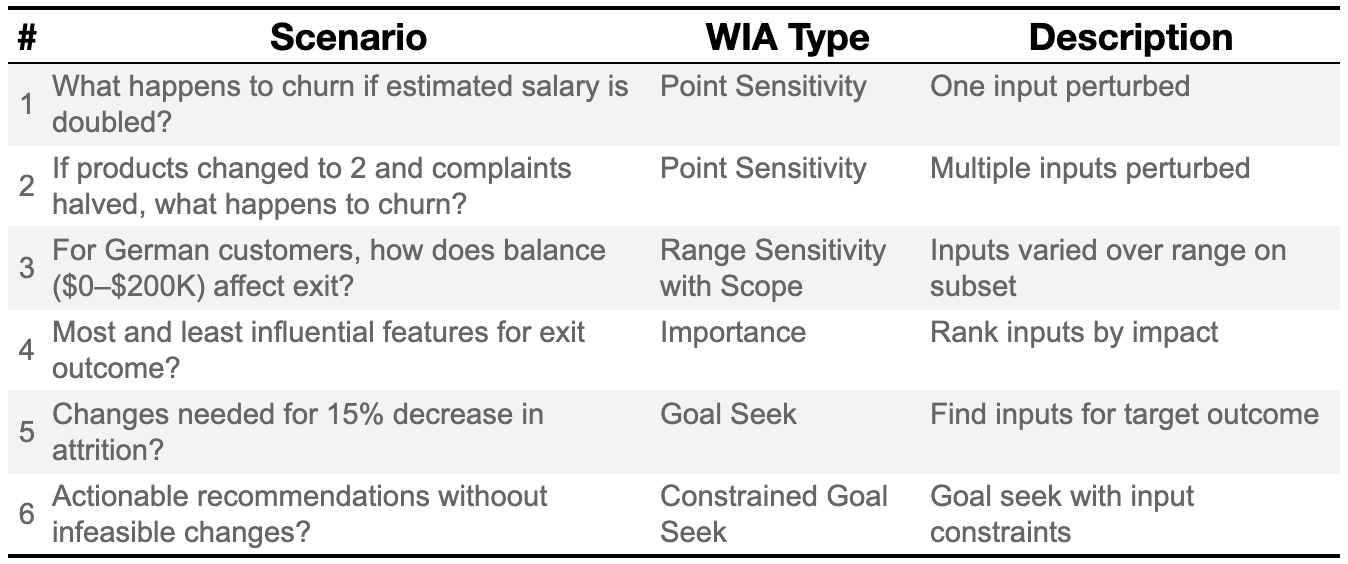}
    \caption{Six WIA scenarios spanning different analysis types used in our formative exploration of spreadsheet and BI tools.}
    \label{fig:exploration_scenarios}
    \Description{A table listing the six what-if analysis scenarios used in the formative exploration, one per row. Each row gives a scenario number, the scenario description, its WIA type, and a short characterization. The scenarios span the main analysis types: two point-sensitivity scenarios (one input perturbed, and multiple inputs perturbed), a range-sensitivity-with-scope scenario (inputs varied over a range on a subset), an importance scenario (ranking inputs by impact), a goal-seek scenario (finding inputs for a target outcome), and a constrained goal-seek scenario (goal-seek with input constraints), all posed over the Bank Customer Attrition dataset.}
\end{figure}

To explore the WIA capabilities of the tools, six WIA scenarios were performed in each.
These scenarios are listed in \autoref{fig:exploration_scenarios}.

\subsubsection{Procedure}
Three authors independently attempted all six WIA scenarios in each of the four tools.
We imported the dataset and explored each tool's WIA capabilities to answer the scenarios, consulting documentation, blogs, and tutorials to overcome tool-specific learning barriers.
Each author spent approximately 4.5 hours, with sessions recorded and jointly reviewed.

\subsection{Benchmark Details}
\label{sec:supp_benchmark}
This section provides additional detail on the datasets, construction quality metrics, ground-truth reconciliation, and PSL encoding of the benchmark introduced in Section 3.

\begin{figure*}[!b]
    \centering
    \includegraphics[width=\textwidth]{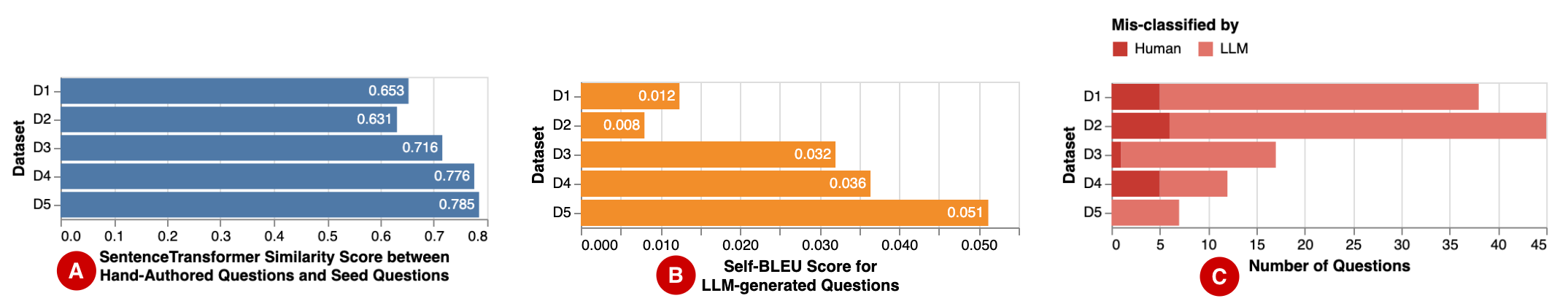}
    \caption{Benchmark construction analysis. (A) SentenceTransformer similarity between seed and hand-authored questions across datasets, indicating preservation of linguistic structure. (B) Self-BLEU scores for LLM-generated questions, indicating high diversity in phrasing. (C) Distribution of WIA-type interpretations by humans and LLMs across datasets.}
    \label{fig:similarity}
    \Description{Three horizontal bar charts, one per panel, with the five datasets D1 to D5 on the vertical axis. Panel A shows SentenceTransformer similarity between seed and hand-authored questions; bars range from about 0.63 to 0.79 across datasets, indicating that adapted questions preserved the structure of their seed questions. Panel B shows self-BLEU scores for the LLM-generated questions; bars are very low, from about 0.008 to 0.051, indicating high diversity in phrasing. Panel C shows the number of WIA-type interpretation mismatches by humans and by LLMs across datasets, with two bars per dataset; LLM mismatch counts are generally higher than human counts, and both vary across datasets.}
\end{figure*}

\subsubsection{Datasets}
We selected five publicly available datasets from Kaggle and the UCI Machine Learning Repository, representing distinct decision-making contexts where WIA is valuable.

\textbf{D1. Bank Customer Attrition}~\cite{bankcustomerattritiondataset}. The same dataset used in our formative exploration of existing tools. It supports what-if questions around customer churn at a bank, with demographic attributes (e.g., \textit{Geography}), account-related features (e.g., \textit{CreditScore, Tenure}), service feedback (e.g., \textit{Satisfaction Score, Points Earned}), and a binary churn outcome (\textit{Exited}).

\textbf{D2. Email Campaign Management}~\cite{emailcampaigndataset}. Supports what-if questions around what influences whether customers open or ignore campaign emails, with campaign attributes (e.g., \textit{Email Campaign Type, Email Source Type}), content features (e.g., \textit{Word Count, Total Links, Total Images}), contextual information (e.g., \textit{Customer Location, Time Email Sent Category}), and a binary open/ignore outcome (\textit{Email Status}).

\textbf{D3. Media Spend and Sales Attribution}~\cite{mediaspendsdataset}. Supports what-if questions around how reallocating spend or impressions across channels impacts sales, with temporal attributes (e.g., \textit{Calendar Week}), channel-level exposure features (e.g., \textit{Google Impressions, Email Impressions}), aggregate engagement (\textit{Overall Views}), and the business outcome (\textit{Sales}).

\textbf{D4. Marketing Campaign Response}~\cite{marketinganalyticsdataset}. Supports analysis of customer responses to direct marketing campaigns, with socio-demographic attributes (e.g., \textit{Income, Age}), purchasing behavior (e.g., \textit{MntWines, MntFruits}), campaign interaction features (e.g., \textit{NumDealsPurchases, NumWebPurchases, NumCatalogPurchases}), campaign acceptance indicators (\textit{AcceptedCmp1--AcceptedCmp5}), service feedback (\textit{Complain}), and a binary success outcome (\textit{Response}).

\textbf{D5. Spotify Revenue, Expenses and Its Premium}~\cite{spotifyrevenuedataset}. Captures business-level financial and subscription performance over time, supporting what-if questions around profitability, pricing, user adoption, and marketing/R\&D investment. It includes temporal attributes (\textit{Date}), financial indicators (e.g., \textit{Total Revenue, Gross Profit}), premium-specific metrics (e.g., \textit{Premium Revenue, Premium Gross Profit}), advertising-related metrics (e.g., \textit{Ad Revenue, Ad Gross Profit}), user engagement (e.g., \textit{MAUs}), and operational expenses (e.g., \textit{Sales and Marketing Cost}).

\subsubsection{Construction Quality Metrics}
We report the construction-quality checks summarized in Section 3.
To confirm that hand-authored questions preserved the structure of the seed questions they were adapted from, we measured SentenceTransformer similarity between each adapted question and its seed; similarity ranged from 0.63 to 0.79 across datasets (\autoref{fig:similarity}A), indicating consistent structure.
To confirm that LLM-generated questions were not repetitive, we measured self-BLEU; low scores (0.008--0.051; \autoref{fig:similarity}B) indicate high diversity in phrasing.

\subsubsection{Ground-Truth Reconciliation}
Two coders independently annotated all 405 questions on the six intent components (Section 3.4) and reconciled discrepancies through discussion.
Of the 179 questions with initial disagreement, the majority stemmed from interpretive ambiguity in the NL question rather than annotation error--most commonly, whether a question implied a scoped or full-dataset analysis, and whether a continuous change implied point or range sensitivity.
For each such case we record the source of disagreement and the rationale for the reconciled intent; 19 cases that could not be resolved by the two coders were adjudicated by a third.

\subsubsection{Encoding Ground-Truth Intent as Intermediate Representation via \psl}
For the workflow, each question's six-component ground-truth intent is additionally encoded as a \psl specification using the schema.
Encoding surfaced schema refinements--for example, adding a \codeN{top} property for importance questions (e.g., \codeN{"top": 1} for ``What is the most salient feature?'').
The reconciled intent for each question determined its canonical specification.

\subsection{Direct-Generation Experiment: Details}
\label{sec:supp_direct}
This section provides the full setup, scoring protocol, validation, and additional qualitative examples for the motivating experiment in Section 4.

\subsubsection{Generation Setup and Prompt Hardening}
For each question, the model is given the NL question and a schema summary of the dataset (column names and types), and asked to emit a single self-contained Python script that (i) trains a scikit-learn RandomForest--a classifier for binary outcomes, a regressor for continuous ones---and (ii) emits an interactive React component with Material UI controls (sliders for continuous variables, selectors for categorical ones, labeled badges for scope) bound to a linked Vega-Lite visualization and a live prediction readout.
This matches the rendering stack and three-part output contract our workflow compiles to (Section 5); the sole difference is that here the interface is authored directly by the LLM rather than compiled from a specification.

To ensure the experiment tests analytical understanding rather than incidental coding robustness, we hardened the generation prompt on pilot questions (disjoint from the evaluation sample) against implementation pitfalls unrelated to analytical intent: string-escaping collisions between generated JSX and Python, Vega-Lite data-binding conventions, and numeric-type serialization between the model and the front end.
Guardrails addressing these were added until pilot generations were free of such errors, at which point the prompt was frozen and left unchanged for the reported evaluation.
The full prompt is included on our OSF supplementary materials~\cite{supplementary}.

\subsubsection{Sample}
From the 405-question benchmark (Section 3) we drew, per model, a stratified random sample of 150 questions using a fixed seed, allocating proportionally across the 5 datasets and, within each dataset, across its labeled WIA types, so the sample mirrors the composition of the full benchmark.
All three models were evaluated on the same sampled questions.
The formative experiment uses \texttt{Claude-Sonnet-4.5}, whereas the workflow evaluation (Section 4) uses \texttt{Claude-Sonnet-4}, a snapshot in the same model family; provider snapshots are periodically retired and superseded.
The two evaluations are reported independently and never compared to one another, so this difference does not affect any result.

\subsubsection{Scoring Protocol}
Each generated interface is scored on the six representation-neutral components of analytical intent defined in Section 3.4: scope, outcome variable, perturbation, constraints, objective, and analysis type.
An LLM judge receives the generated code, its execution output (or, if execution failed, the resulting error), and the ground-truth intent, and returns for each component a judgment of match, mismatch, or not-applicable together with a one-line justification.
A component is marked not-applicable only when the ground truth has no such field for that question (e.g., an unscoped question has no scope component); such cases are excluded from that component's denominator.
Because a crashed generation's code still reveals the analysis the model attempted, non-rendering interfaces are scored on the same basis and are included in the component rates.

\subsubsection{Validation Against Human Judgment}
To assess the reliability of the automated scoring, one author independently hand-scored a stratified random sample of 40 questions from the 150 questions used in this experiment.
For each question, the generated code and interface were evaluated against the same ground truth and scoring rubric, without access to the automated judgments.
Across the resulting 240 component judgments (40 interfaces $\times$ 6 components), the human and automated scores agreed on 216 cases (90.0\%, Cohen's $\kappa = 0.847$).
Agreement was 90.0\% on the random subsample alone, the curated subset alone, and the two combined, indicating the curated cases did not inflate the estimate.
Per-component agreement was highest for outcome variable (100\%) and scope (97.5\%) and lowest for objective (80.0\%, $\kappa = 0.678$); we accordingly read the objective rates in the main text with somewhat greater caution than the other five components.

\begin{table}[t]
\centering
\caption{Applicable $N$ (match + mismatch) per component per model, i.e., the number of sampled questions (out of 150) whose ground truth specifies that component. Rates in the main text are computed over these denominators.}
\label{tab:supp_applicable_n}
\begin{tabular}{lcccccc}
\toprule
Model & Scope & Outcome Var. & Perturbation & Constraints & Objective & Analysis Type \\
\midrule
GPT-5 & 78 & 150 & 123 & 46 & 102 & 150 \\
GPT-4o & 80 & 150 & 131 & 55 & 131 & 150 \\
Claude-Sonnet-4.5 & 85 & 150 & 126 & 43 & 80 & 150 \\
\bottomrule
\end{tabular}
\end{table}

\subsubsection{Additional Qualitative Examples}
The main paper (Section 4) presents one worked example--a constrained goal-seek that renders fully yet reports an achievable target as impossible.
Here we detail four further examples spanning the other functional error types, including a case where the same scope confusion produces an outright crash rather than a silent error.

\textbf{Scope silently re-defined (importance with scope).}
For \textit{``During calendar weeks with high promotional activity, assess the comparative significance of Affiliate vs.\ Paid Views in propelling Sales,''} the intended scope is high-\textit{sales} weeks (Sales above the third quartile). 
Claude-Sonnet-4.5 instead interpreted ``high promotional activity'' as high promotional \textit{impressions}, scoping to weeks where combined Google, Facebook, and Affiliate impressions exceed their third quartile--a different, though plausible, set of weeks.
The generated interface is otherwise correct--it identifies the right outcome, performs the requested comparison, and even labels its scope prominently (``High Promotional Activity Weeks: Total Promo Impressions $\geq$ 1{,}340{,}852'').
But that label reflects the model's own interpretation, silently substituted for the intended one; a user cannot tell that the entire comparison rests on a different subset of weeks than they asked about.
The scope choice exists only in the generated code, so the substitution is undetectable from the interface.

\textbf{Objective neglect (goal-seek answered as a sweep).}
For \textit{``Determine the level of MntSweetProducts and NumDealsPurchases needed for customers with PhDs to reach a Response rate of 85\%,''} the ground-truth objective is to \textit{maximize} response.
GPT-5's generated code instead sweeps the two inputs and reports where predicted response crosses the fixed 85\% figure, with no optimization step anywhere in the generated implementation.
The rendered interface exposes sliders and a prediction readout and appears to answer the question; the absence of any objective is visible only by reading the computation itself.

\textbf{Outcome-variable substitution (same error, two models).}
For \textit{``Considering constraints that Income should not surpass \$75,000 and NumCatalogPurchases must remain under 8, what are optimal settings for other spending categories?,''} the ground truth specifies \texttt{Response} as the outcome to maximize.
Both GPT-4o and Claude-Sonnet-4.5 independently optimized \texttt{MntTotal} (total spending) instead; GPT-5, given the identical question, optimized \texttt{Response} correctly.
Two of three models substituting the same wrong target on the same question indicates the error is a property of the task as posed rather than one model's slip: nothing in a one-shot generation contract forces the model to name and commit to the variable it is actually optimizing.

\textbf{Constraints silently dropped (same error, two models).}
For \textit{``To reduce the risk of customer attrition, what actionable recommendations can be made without suggesting infeasible changes, such as drastically increasing the customer's age or reducing their tenure?,''} the ground truth bounds any recommended change to Age within $\pm5$ years and Tenure within $\pm3$ years.
Both GPT-5 and GPT-4o generated code applying no such bounds, leaving Age and Tenure unconstrained.
The bound was stated only in NL and had no forced representation in the generated code, so it was dropped entirely while the resulting recommendations still appeared reasonable.

\subsection{Workflow Evaluation: Details}
\label{sec:supp_evaluation}
This section provides additional detail for the workflow evaluation in Section 6: the WIA-type classification step and its agreement with a human baseline, the per-dataset distribution of generation errors, and the two automated error-detection experiments.

\subsubsection{WIA-Type Classification}
\label{sec:supp_classification}
Before generating a \psl specification, each LLM classifies a question into one of the 11 WIA types (Section 6.1).
Each model was prompted with dataset context, type definitions, and a few examples; to reduce ordering effects we randomized the presentation order of the types, collected three predictions per question, and took the majority vote.
Three coders independently annotated the WIA type of every question to serve as a human baseline.

Across the 405 questions, we observed 119 questions on which the majority LLM label disagreed with the human label.
On inspection, these fell into two groups.
The large majority (102 of 119) were genuine LLM misclassifications: the model over-interpreted phrasing (e.g., reading an incidental mention of a subgroup as a scoping requirement) or failed to distinguish scoped from unscoped variants of the same analysis type.
The remaining 17 were cases where the LLM label was in fact more appropriate than the human label--most often continuous-range questions that a coder had labeled as point sensitivity (T1) but that are more precisely range sensitivity (T3).
We reviewed all 119 cases and adopted the more accurate label before specification generation, so that generation errors downstream are not confounded by upstream type-classification disagreements.

\subsubsection{Per-Dataset Distribution of Generation Errors}
\label{sec:supp_perdataset}
\begin{figure}[t]
    \centering
    \includegraphics[width=\columnwidth]{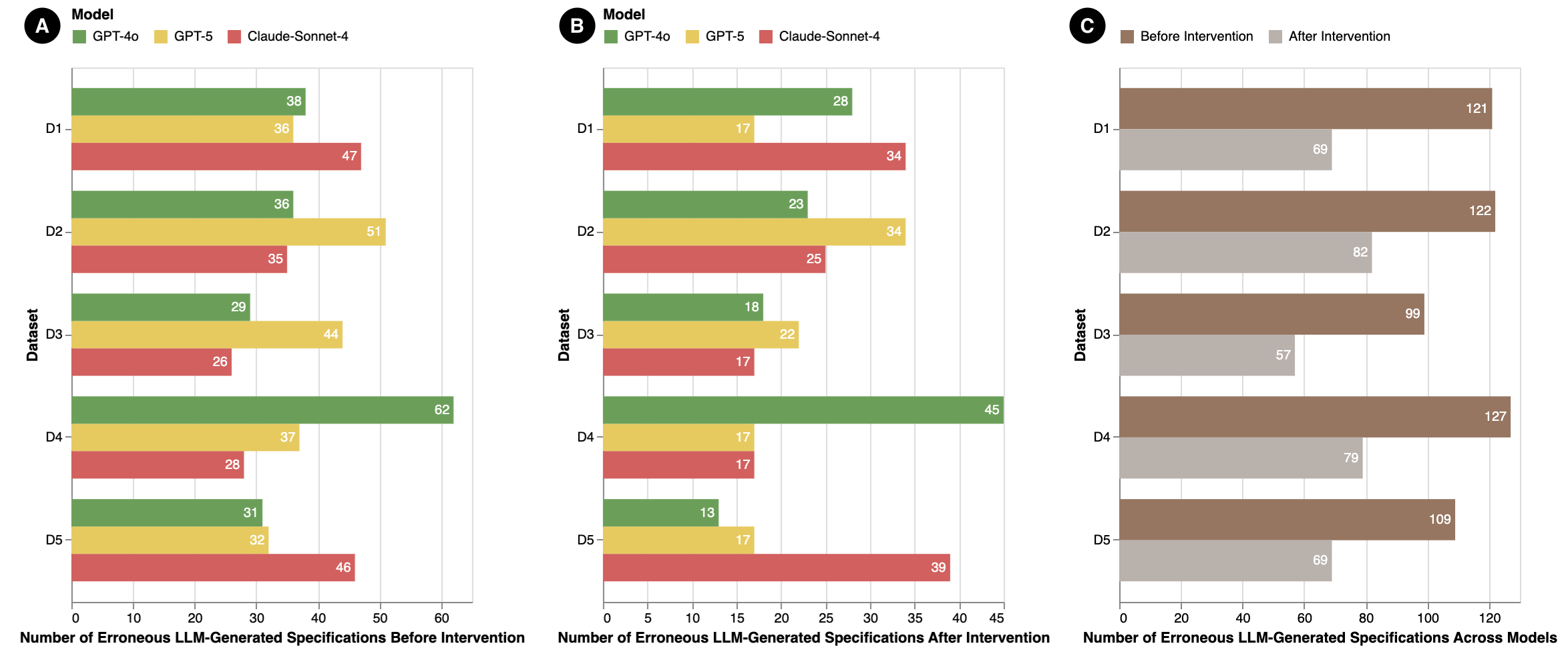}
    \caption{Number of erroneous LLM-generated specifications per dataset and model, before and after targeted repair.}
    \label{fig:supp_perdataset}
    \Description{Grouped horizontal bar charts showing the number of erroneous LLM-generated specifications for each of the five datasets D1 to D5, with separate bars for GPT-4o, GPT-5, and Claude-Sonnet-4, shown before and after targeted repair. Before repair, per model-dataset counts range from about 26 to 62, with the highest counts for GPT-4o on D4 and GPT-5 on D2, and comparatively high counts for Claude-Sonnet-4 on D1 and D5. After repair, every bar shortens, with counts falling to roughly the 13 to 45 range; aggregated across models, total errors per dataset drop from between 99 and 127 before repair to between 57 and 82 after, a consistent reduction across all datasets.}
\end{figure}

\autoref{fig:supp_perdataset} reports the number of erroneous specifications per dataset per model, before and after targeted repair.
Error counts before intervention range from 26 to 62 per model--dataset cell, with higher counts for GPT-4o (e.g., 62 on D4) and GPT-5 (e.g., 51 on D2) and comparatively higher counts for Claude-Sonnet-4 on D1 (47) and D5 (46).
After intervention, counts decrease across all datasets and models, typically to the 13--45 range (e.g., D4 falls from 62 to 45 for GPT-4o, and from 28 to 17 for Claude-Sonnet-4).
Aggregated across models, total errors per dataset fall from 99--127 before intervention (e.g., 127 on D4, 122 on D2) to 57--82 after (e.g., 79 on D4, 82 on D2), a consistent reduction across datasets.

\subsubsection{Automated Error-Detection Experiments}
\label{sec:supp_detection}
We report the two error-detection experiments summarized in Section 6.3.
Both ask whether an LLM can identify errors in a generated specification; both compare LLM judgments against human annotators.

\bpstart{Experiment A: Binary detection}
\textit{Task.} Given a WIA question and its LLM-generated \psl specification, the model predicts whether \textit{any} error is present (yes/no), without identifying a category.

\textit{Setup.} Each LLM receives the question, dataset context, the \psl schema, and the generated specification, and outputs a binary label (1 = error, 0 = no error).
Three human annotators independently label the same specifications, with majority vote as reference.
Evaluation covers all 405 questions per model.

\begin{figure}[t]
    \centering
    \includegraphics[width=\columnwidth]{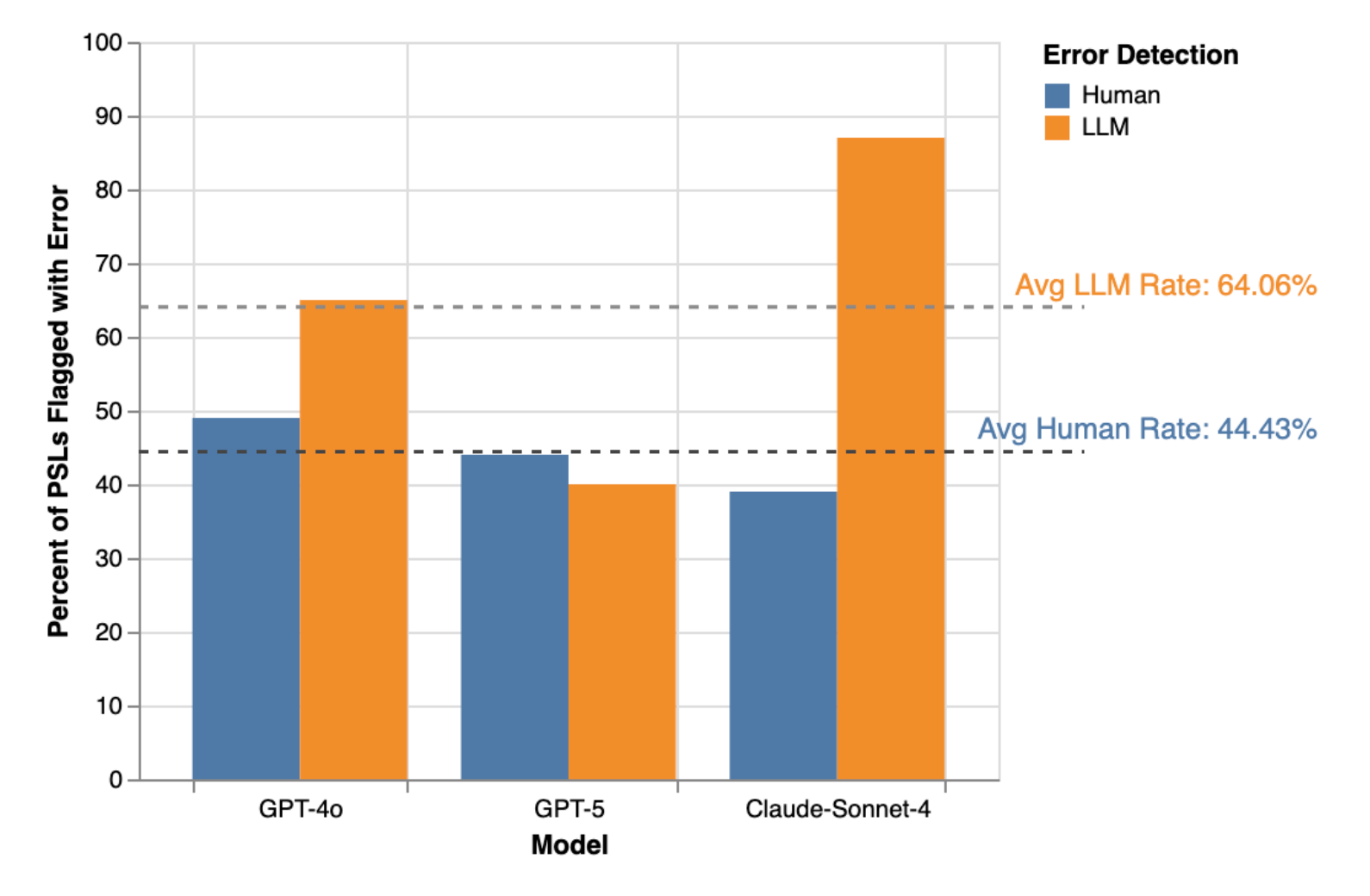}
    \caption{Experiment A: binary detection of any error in LLM-generated specifications, by model, against the human flagging rate.}
    \label{fig:supp_detection}
    \Description{A bar chart showing, for each of the three models (GPT-4o, GPT-5, and Claude-Sonnet-4), the percentage of specifications the model flagged as containing an error in the binary error-detection experiment. Two reference lines mark the average LLM flagging rate of about 64 percent and the average human flagging rate of about 44 percent. GPT-4o and Claude-Sonnet-4 flag errors above the human rate, indicating over-detection, while GPT-5 flags at a rate closer to or slightly below the human baseline. Overall the models flag errors more often than human annotators do.}
\end{figure}

\textit{Results.} Across models, accuracy is 64.06\% (TP\,=\,145, TN\,=\,103, FP\,=\,87, FN\,=\,70; precision\,=\,0.625, recall\,=\,0.674, F1\,=\,0.649), with modest agreement ($\kappa = 0.218$).
LLMs flag 64.06\% of specifications as erroneous versus 44.43\% flagged by humans, indicating systematic over-detection.
Model-wise, GPT-4o and Claude-Sonnet-4 over-estimate errors while GPT-5 slightly under-estimates them (\autoref{fig:supp_detection}).

\bpstart{Experiment B: Per-category diagnosis}
\textit{Task.} For each specification, the model predicts which error categories (EC1--EC9) are present.

\textit{Setup.} Inputs include the question, dataset context, the \psl schema, the generated specification, and, for each category, a definition bundle with 2--3 positive and negative examples.
For each category the model outputs a binary present/absent decision; human annotators provide reference labels under the same protocol.
To balance coverage and cost we sampled 140 questions and evaluated all three models, yielding 420 specifications, each model evaluating its own outputs.

\begin{figure*}[t]
    \centering
    \includegraphics[width=\textwidth]{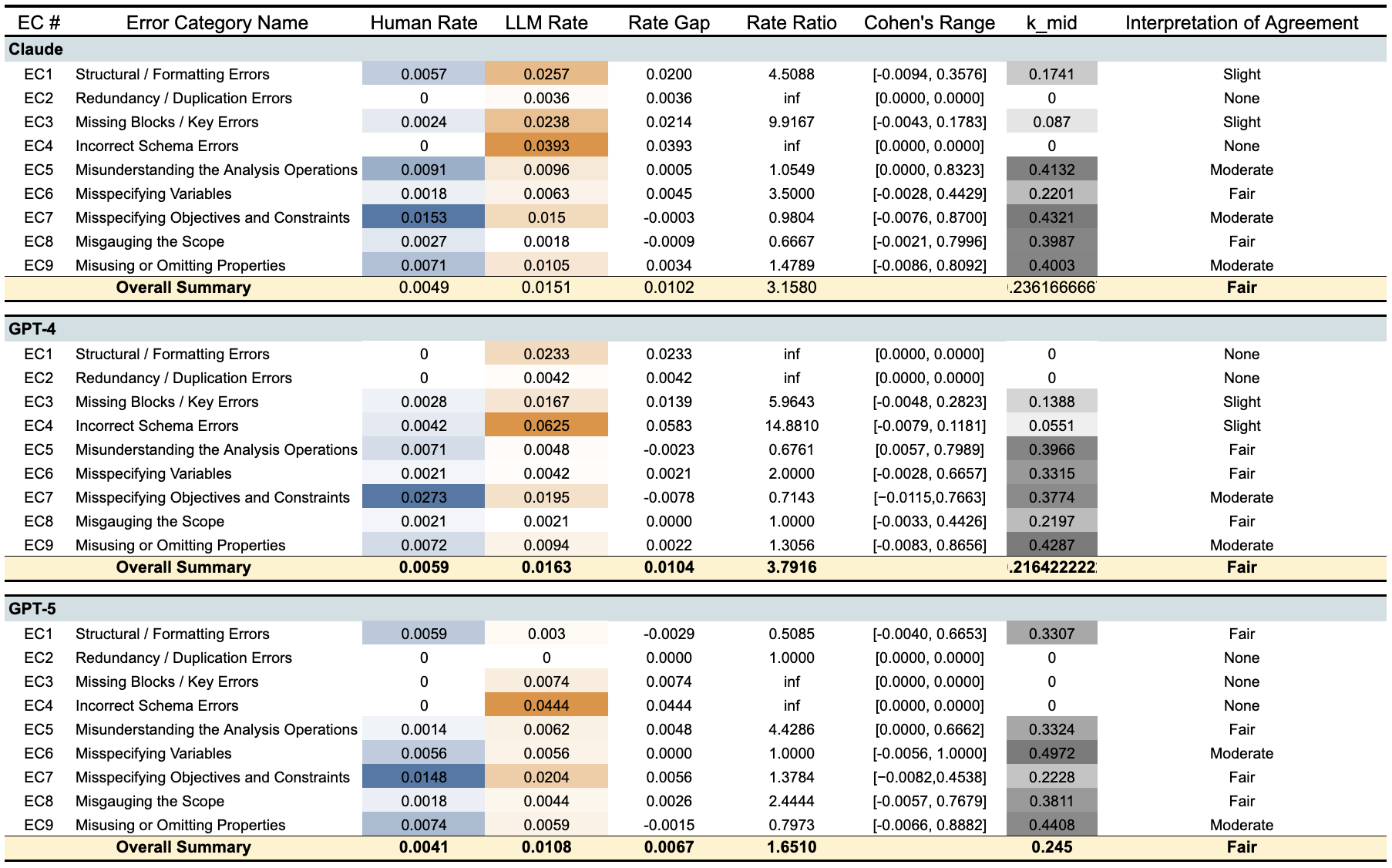}
    \caption{Experiment B: per-category agreement between human annotators and LLMs (EC1--EC9). For each category we report the human positive rate, the LLM positive rate, the rate gap (LLM$-$human, percentage points), the rate ratio (LLM/human), and Cohen's $\kappa$. Higher gap and ratio values indicate greater LLM over-estimation of errors.}
    \label{fig:supp_detection_percategory}
    \Description{A table reporting per-category agreement between human annotators and LLMs across the nine error categories EC1 to EC9, with sections for each model (GPT-4o, GPT-5, and Claude). For each error category the table gives the human positive rate (fraction of specifications humans flagged as having that error), the LLM positive rate, the rate gap (LLM minus human, in percentage points), the rate ratio (LLM divided by human), and Cohen's kappa with an interpretation label (none, slight, fair, or moderate agreement). Non-functional categories (EC1 to EC4) show none-to-slight agreement and large gaps and ratios, indicating substantial LLM over-flagging, while functional categories (EC5 to EC9) show fair-to-moderate agreement and smaller gaps; across the table LLM positive rates exceed human rates, with over-flagging ranging from roughly three to over twenty times the human rate.}
\end{figure*}

\textit{Results.} Agreement is low ($\kappa \approx 0.23$--0.26), with LLMs over-flagging errors at 3--23$\times$ the human rate.
Non-functional errors (EC1--EC4) show none-to-slight agreement, reflecting over-sensitivity to structural issues; functional errors (EC5--EC9) show fair-to-moderate agreement, indicating better calibration on semantic interpretation.
Per-category human and LLM positive rates, rate gaps and ratios, and Cohen's $\kappa$ are reported in \autoref{fig:supp_detection_percategory}.
Overall, LLMs surface potential errors but cannot reliably categorize them--the basis for the human-in-the-loop repair strategy in Section 6.3.

\end{document}